# IndicFairFace: Balanced Indian Face Dataset for Auditing and Mitigating Geographical Bias in Vision-Language Models

**Aarish Shah Mohsin**
Aligarh Muslim University, India
aarishshahmohsin50@gmail.com

**Mohammed Tayyab Ilyas Khan**
Aligarh Muslim University, India
tayyabilyas963@gmail.com

**Mohammad Nadeem**
Aligarh Muslim University, India
mnadeem.cs@amu.ac.in

**Shahab Saquib Sohail**
Jamia Hamdard, India
shahabssohail@jamiahamdard.ac.in

**Erik Cambria**
Nanyang Technological University, Singapore
cambria@ntu.edu.sg

**Jiechao Gao**
Stanford University, USA
jiechao@stanford.edu

February 13, 2026

## Abstract

Vision-Language Models (VLMs) are known to inherit and amplify societal biases from their web-scale training data with Indian being particularly misrepresented. Existing fairness-aware datasets have significantly improved demographic balance across global race and gender groups, yet they continue to treat "Indian" as a single monolithic category. The oversimplification ignores the vast intra-national diversity across India's 28 states and 8 Union Territories and leads to representational and geographical bias. To address the limitation, we present IndicFairFace, a novel and balanced face dataset comprising 14,400 images representing India's geographical diversity. Images were sourced ethically from Wikimedia Commons and open-license web repositories and uniformly balanced across states and gender. Using IndicFairFace, we quantify intra-national geographical bias in prominent CLIP-based VLMs and reduce it using post-hoc Iterative Nullspace Projection debiasing approach. We also show that the adopted debiasing approach does not adversely impact the existing embedding space as the average drop in retrieval accuracy on benchmark datasets is less than 1.5%. Our work establishes IndicFairFace as the first benchmark to study geographical bias in VLMs for the Indian context. The dataset and code can be found here: https://github.com/aarishshahmohsin/IndicFairFace

## 1 Introduction

Vision-Language Models (VLMs), whether generative or discriminative, have advanced multimodal understanding [1]. By jointly embedding text and image representations, they achieved state-of-the-art zero-shot transfer across classification, retrieval, and captioning tasks [2, 3]. The success of VLMs led to their wide adoption in various applications [4]. However, several studies show that VLMs encode systematic social and geographic biases inherited from the image–text data used during pre-training [5, 6, 7, 1]. The biases lead to allocation and representation harms when models associate negative stereotypes with certain demographic or cultural groups[8, 5].

Though fairness research has expanded in recent years and the benchmarks such as FairFace [9], UTKFace [10], CelebA [11] etc. have improved racial balance, their scope remains largely Western-centric [12]. Additionally, the global race taxonomy fails to preserve intra-national diversity in VLMs, particularly for countries like India where rich facial variation is often collapsed into a few broad categories [13]. As a result, the Indian facial representation in VLMs is treated as monolithic where embeddings often over-represent a few urban regions while ignoring others to create a distorted notion of 'Indianness' (see Figure 1). The imbalance poses ethical and technical risks when VLMs are deployed in systems for identity verification, recruitment or recommendation where equitable representation is critical.

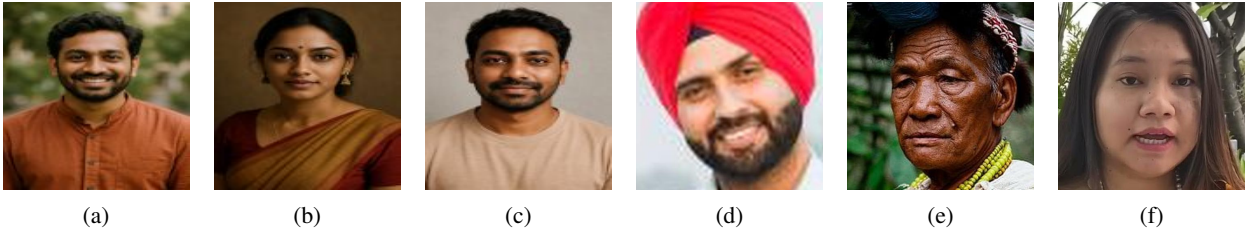

| (a) | (b) | (c) | (d) | (e) | (f) |

Figure 1: When prompted about Indian person (1a), Indian female (1b) and Indian male (1c), GPT-5 consistently generates images representing people only a few regions/states while ignoring other parts of the country. The diversity of faces in India is shown in (1d,1e,1f.

Table 1: Comparison of fairness-aware face image datasets across key characteristics. (✓= Yes / present, ✗= No / absent, # = Partially present or unclear)

| Sl. | Dataset | References/Year | #Images / Subjects | In-the-wild | Balanced (Gender & Race / State) | Demographic Scope | Indian context | Type (Face / Full Body) |
|---|---|---|---|---|---|---|---|---|
| 1 | CelebA | [11]/2015 | 200K / 10K | ✓ | ✗ | Mostly White, Western | ✗ | Face |
| 2 | LFW | [14]/2008 | 13K / 5K | ✓ | ✗ | Western celebrity faces | ✗ | Face |
| 3 | UTKFace | [15]/2017 | 20K | ✓ | # | 4 Races (merged) | ✗ | Face |
| 4 | DiF | [16]/2019 | 1M | ✓ | ✗ | Derived from Flickr-YFCC100M | ✗ | Face |
| 5 | FairFace | [17]/2021 | 108K | ✓ | ✓ | 7 race groups | ✗ | Face |
| 6 | PPB | [18]/2018 | 1K | ✗ | ✗ | Gov. officials (controlled lighting) | ✗ | Face |
| 7 | Cas. Conv. v2 | [19]/2022 | 45K / 3K | ✗ | # | Global consented users | ✗ | Face |
| 8 | Deity-TU Face | [20]/2014 | 49.8K / 524 | ✓ | ✗ | North-East India (single region) | # | Face |
| 9 | FHIBE | [21]/2025 | 10K / 1.9K | ✓ | ✓ | 81 countries, 5 continents | # | Face + Full body |
| 10 | IndicFairFace | (Proposed) | 14.4K | ✓ | ✓ | 28 States + 8 UTs, India | ✓ | Face |

Addressing this challenge requires geographically balanced datasets that capture India's internal diversity. However, to the best of our knowledge, no existing Indian public face dataset provides state-wise balance. Therefore, we introduce IndicFairFace, the first state-balanced facial dataset for auditing and mitigating geographical bias in VLMs for the Indian context. It contains 14,400 public-domain face images, equally distributed across India's 28 states and 8 Union Territories with gender parity. Images were sourced through Wikimedia Commons (SPARQL queries) and SerpAPI-based web retrieval. The dataset can be used for multiple purposes such as 1) auditing pre-trained VLMs for regional bias, 2) training region-aware or debiased embeddings and, 3) evaluating fairness of retrieval or captioning systems.

To futher study geographical bias for Indian context, we used text-to-image retrieval VLMs (such as CLIP [2], Open-CLIP [22], SigLIP [23], MetaCLIP [24], MobileCLIP[25], OpenVision[26] and their variants). Primarily, we looked into the distribution of states in the outputs when VLMs were prompted with the textual prompt containing 'Indian'. The findings show clear signs of geographical bias where a few states/regions dominate the distribution. To mitigate the bias, we employed Iterative Nullspace Projection (INLP) approach to remove learned bias directions from the CLIP feature space, combined with Spherical Linear Interpolation (SLERP) for controlled projection strength and similarity compensation to preserve semantic alignment. The adopted debiasing approach maintains model utility while enforcing embedding-level fairness. Empirically, the debiased model reduced inter-state variance in mean similarity scores by ≈37% while incurring <1.5% loss in retrieval accuracy on external benchmark datasets.

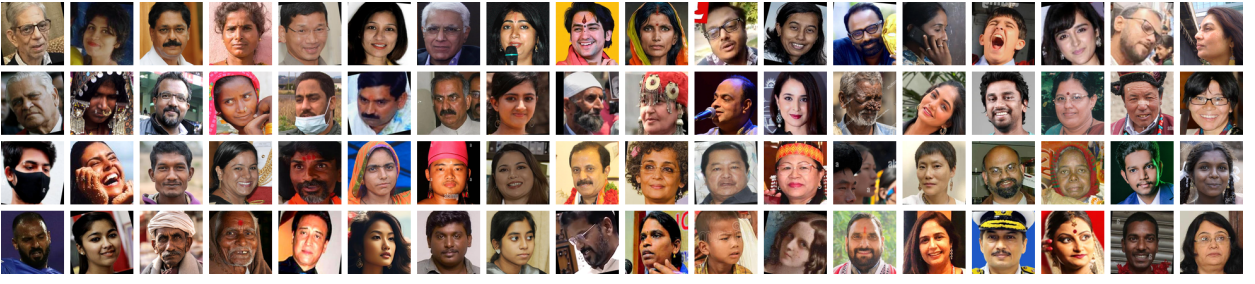

Figure 2: A sample of male and female faces for all provincial regions in India.

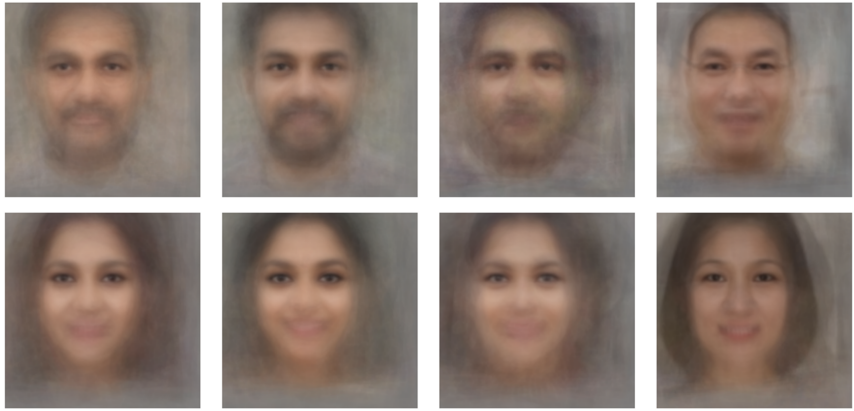

Figure 3: The average faces of male (top row) and female (bottom row) for four prominent regions: North, South, West and North-East (left to right), by superimposing the images of individuals from that region.

## 2 Related Work

Several datasets have been introduced to support algorithmic fairness research in facial analysis. Among them, **FairFace** [9] is particularly notable for its large, demographically balanced collection of face images spanning seven racial categories, gender, and age. Other widely used datasets include UTKFace [10] for age and gender estimation, as well as older benchmarks such as Labeled Faces in the Wild (LFW) [27] and CelebA [28]. While invaluable, these datasets share a critical limitation, i.e., they treat "Indian" or "South Asian" as a single, monolithic group. More recent, consent-driven datasets such as Facebook's **Casual Conversations** [29] are used to measure bias but are not focused on the specific cultural or geographical granularity of the Indian context.
A few benchmarks for Indian context exist but most of them are text-based. IndicBias [12], IndiBias [30] and IndianBhED [31] have created detailed taxonomies of harmful stereotypes and confirm that **region, religion, and caste** are primary axes of bias in India, alongside gender. The SPICE dataset [32] also collects and analyzes stereotypes pertaining to state, religious, and ethnic identities. However, Indian face-based datasets are still limited. DeitY-TU face dataset [20] and IISC face detection (IISCIFD) dataset [33] are small, old, captured in constrained settings and focus on specific regions. Moreover, most publicly available Indian datasets are skewed towards celebrities and urban populations [34].Recently, several large-scale Indian-centric datasets have been introduced, including Chitrarth [35] (a VLM benchmark for India), Chitrakshara [36] (a large multilingual multimodal dataset), and Chitranuvad [37] (for multimodal translation). However, all these datasets are linguistically oriented and do not contribute to face-centric research. Similarly, FolkTalent [38] (focused on Indian folk paintings) and FKG.in [39] (a knowledge graph for Indian food) are Indian-centric but not related to the facial domain, nor do they emphasize a balanced, state-wise representation of Indian faces. To that end, the proposed **IndicFairFace** dataset introduces a novel contribution by offering a face-centric, demographically diverse collection that achieves both state-wise and gender balance within India (see Table 1).

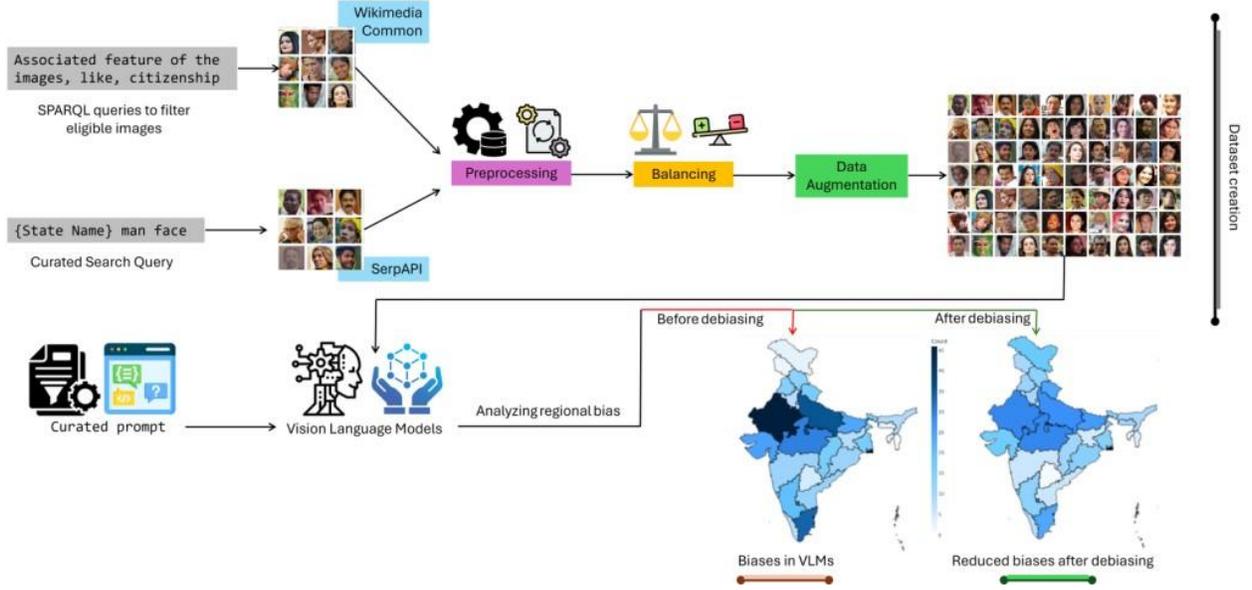

Figure 4: Three tier architecture of the proposed work including dataset creation, bias identification and debiasing in VLMs.

## 3 IndicFairFace dataset

Models trained on Western-centric data often perform poorly on faces from the North-Eastern, coastal and tribal regions of India. Moreover, India-centric datasets are also biased towards metropolitan regions and celebrities [40]. In this section, the curation process of IndicFairFace dataset is elaborated which is balanced and annonated across two demographic dimension i.e. region (state and UT) and gender. The process of data curation is shown in Figure 4.

### 3.1 Data sources and collection

Since face images are copyrighted, we used; 1) Wikimedia Commons, an open-source, structured repositories of images and 2) web scraping using controlled search APIs (others researchers [41, 42, 43] have also used similar approach). For the first source, we accessed the structured semantic database of Wikidata which connects image assets on Wikimedia Commons to rich metadata such as place of birth, citizenship, gender, etc. Using SPARQL queries, we filtered for items satisfying the following conditions:

- Instance of human
- Country of citizenship: India
- Birthplace associated with a State or Union Territory
- Available image under open license

The SPARQL query ensured accurate geographic attribution and linked every face with its state-level origin. All images were downloaded using a custom Python pipeline that automatically verified licenses (CC0, CC-BY, and Public Domain). In states with insufficient Wikimedia coverage (for example, Mizoram, Tripura, Ladakh, and Andaman & Nicobar Islands), additional images were collected using the Google Image Search API (SerpAPI). Following the established approaches (such as C-TRIP [43] and CLDyB [44]), we constructed state-wise gender-specific queries such as "`{State Name} man face`" and "`{State Name} woman face`". Each retrieved image was validated for open license and manually verified for geographical relevance. The resulting corpus was then cleaned and structured following the same directory hierarchy as the Wikimedia subset.

### 3.2 Face Detection, cleaning and preprocessing

To ensure clear and realistic faces, all downloaded images were processed through a cleaning pipeline built on OpenCV and YOLOv8 [45]. The cleaning included several automated heuristics for face detection (using OpenCV's deep neural network face detector), colour validation (grayscale or black-and-white portraits were discarded using colour variance thresholds to maintain consistency in skin-tone representation), texture filtering (low-variance Laplacian filters were used to detect cartoon-like or synthetic faces, which were excluded) and skin region validation (HSV skin-tone masking was applied to ensure the cropped face had sufficient skin-pixel ratio for realistic representation). Each face was then cropped with a $20\%$ padding margin, resized to $512 \times 512$ pixels and saved in RGB format. Figure 2 illustrates a sample of male and female faces generated and Figure 3 shows the average faces of male and female for four prominent regions of India.

### 3.3 Balancing strategy and augmentation

There are two major reasons for geographical bias i.e. representation gaps and imbalance in sample count of the dataset. Therefore, we standardised the dataset to contain exactly **100 faces per gender per state**. In cases where the available samples were fewer, synthetic augmentations were introduced using mild transformations such as random horizontal flips, rotations, hue jitter, and random crops [9, 46]. The balanced dataset thus comprised **7,200 face images** (3,600 male and 3,600 female) across all 28 states and 8 Union Territories. Each image is associated with two attributes i.e. `state` and `gender`.

To further increase the size of dataset, we used controlled colour-based augmentation. Studies have shown that controlled colour-space variation improves model robustness and enhances fairness across skin tones and illumination contexts [47, 48, 49]. In particular, augmentations involving mild changes in hue and brightness encourage models to learn invariant facial representations and reduces the effect of lighting and camera artefacts [50]. Further, small stochastic colour perturbations yield substantial improvement in generalisation without compromising fidelity [47, 48].

Overall, we publish all variations of our dataset i.e. imbalanced and without face extraction, having 7,200 images (balanced) and having 14,400 images (balanced). Table 2 summaries the key attributes of the IndicFairFace dataset. Detailed state-wise distribution of imbalanced dataset from both sources is available in *Supplementary Material*.

### 3.4 Scope

Though we used the dataset for bias auditing and debiased model training, it can also be used for regional representation studies and state-wise or region-wise (north, south, north-east etc. by combining neighboring states/UTs) face retrieval tasks. The uniform state-gender distribution of the dataset allows for evaluating fairness metrics and robustness under controlled augmentations. Moreover, the imbalanced version can be useful for studying data-driven bias propagation and representation gaps in Indian context. Lastly, the dataset can be used to enrich the existing Indian image datasets.

## 4 Bias audit and mitigation

Here we discuss the overall methodology adopted for bias identification and debiasing process (see Figure 4).

### 4.1 VLMs used

We evaluated CLIP [2], OpenCLIP [22], SigLIP [23], MetaCLIP [24], MobileCLIP[25], OpenVision[26] and their variants for bias identification. Most models pair a Vision Transformer (ViT) image encoder with a Transformer text encoder trained to align modalities in a shared embedding space through contrastive learning [51]. CLIP models are trained on 400M image-text pairs using a dual-encoder contrastive objective [2] while OpenCLIP models are larger and trained on LAION-2B, an open-source image-text corpus [52] and others (SigLIP and MetaCLIP) used internally curated datasets. Detailed discussions about their architectures and training datasets used can be found at [2, 22, 23, 24, 25, 26].

Table 2: Summary of IndicFairFace dataset.

| Aspect | Description |
|---|---|
| #Images | 7,200 (3,600 male + 3,600 female) |
| #Images (augment.) | 14,400 (7,200 male + 7,200 female) |
| #Images (Imbal.) | 13,304 (6,367 male + 6,937 female) |
| Geographical units | 36 (28 States + 8 Union Territories) |
| Image Resolution | $512 \times 512$ (RGB) |
| Source | Wikimedia commons, Web (SerpAPI) |
| Annotations | Gender and state |
| Licenses | CC0 / CC-BY / Public Domain |

### 4.2 Prompt Design

We first created three gender-based categories: 1) Neutral, 2) Male, and 3) Female and created multiple similar prompts for each category. For example, we used prompts such as Indian person, Indian citizen, resident of India, Indian face etc for the neutral category. The objective was to analyze the retrieved distribution of states when prompted for 'India' along with specific gender. The list of all the prompts used is given in *Supplementary Material*. It was observed that multiple prompts for specific gender category did not yield significantly different results.

### 4.3 Train-Test Split

IndicFairFace dataset (we used the balanced version with 7,200 images) was divided into two equal, non-overlapping subsets for training and evaluation to maintain balance across both gender and state representation. Each of the 36 geographical units (28 States and 8 Union Territories) contributes an equal number of samples (50 male and 50 female) to both splits. The augmented samples were also assigned uniformly to prevent information leakage between the splits.

### 4.4 Debiasing

We adopted debiasing strategy based on Iterative Nullspace Projection (INLP) [53] to remove state-specific bias directions from the embedding space of VLMs while preserving semantic utility [54, 55]. The method identifies linear classifiers that best separate face embeddings across Indian states and iteratively projects the representations onto the nullspace of those classifiers. Each projection step eliminates residual components encoding regional identity, ensuring that the resulting embeddings are less correlated with geographic cues. We further integrate Spherical Linear Interpolation (SLERP) [56] to regulate projection strength and prevent over-correction which allows fine-grained control of the fairness–utility trade-off.

To compensate for semantic drift, we apply a similarity-preserving adjustment [57, 58] which aligns the debiased embeddings closer to their original concept vectors (e.g., "Indian person"). The final debiased representations are then re-evaluated using cosine similarity between face and text embeddings, ensuring that fairness improvements do not degrade retrieval performance. The information about each step is mentioned in detail in *Supplementary Material*.

### 4.5 Metrics used

We utilized three measures to assess the degree of geographical bias in the outcomes of VLMs: 1) Top-k similarity, 2) Average similarity for each category and 3) Normalized Jensen-Shannon (JS) score. All three have already been used by researchers for bias assessment [5, 59]. Top-k similarity ranks all images of the evaluation set by their cosine similarity for a given target text prompt and selects the top $k$ images. Average similarity per category measures the model's average alignment with a target concept for a specific subgroup. For each demographic category (e.g., an Indian state), we computed the mean similarity between all its image embeddings and the target text embedding (e.g., "An Indian person"). Normalized Jensen-Shannon (JS) score compares the distributions top-k results with the uniform distribution. It is a symmetric, smoothed, and normalized (bounded between 0 and 1) measure of the statistical distance between two probability distributions.

### 4.6 Experimental setup

The experiments were conducted using Google Colab and a local workstation equipped with an NVIDIA RTX 4050 GPU (6 GB VRAM). The full fairness pipeline, including the model fitting, zero-shot misclassification audit, and comparative evaluation of original versus debiased VLMs, was executed in the same environment. The setup provided sufficient computational resources to handle image-text embeddings while maintaining efficient runtime for debiasing and evaluation tasks.

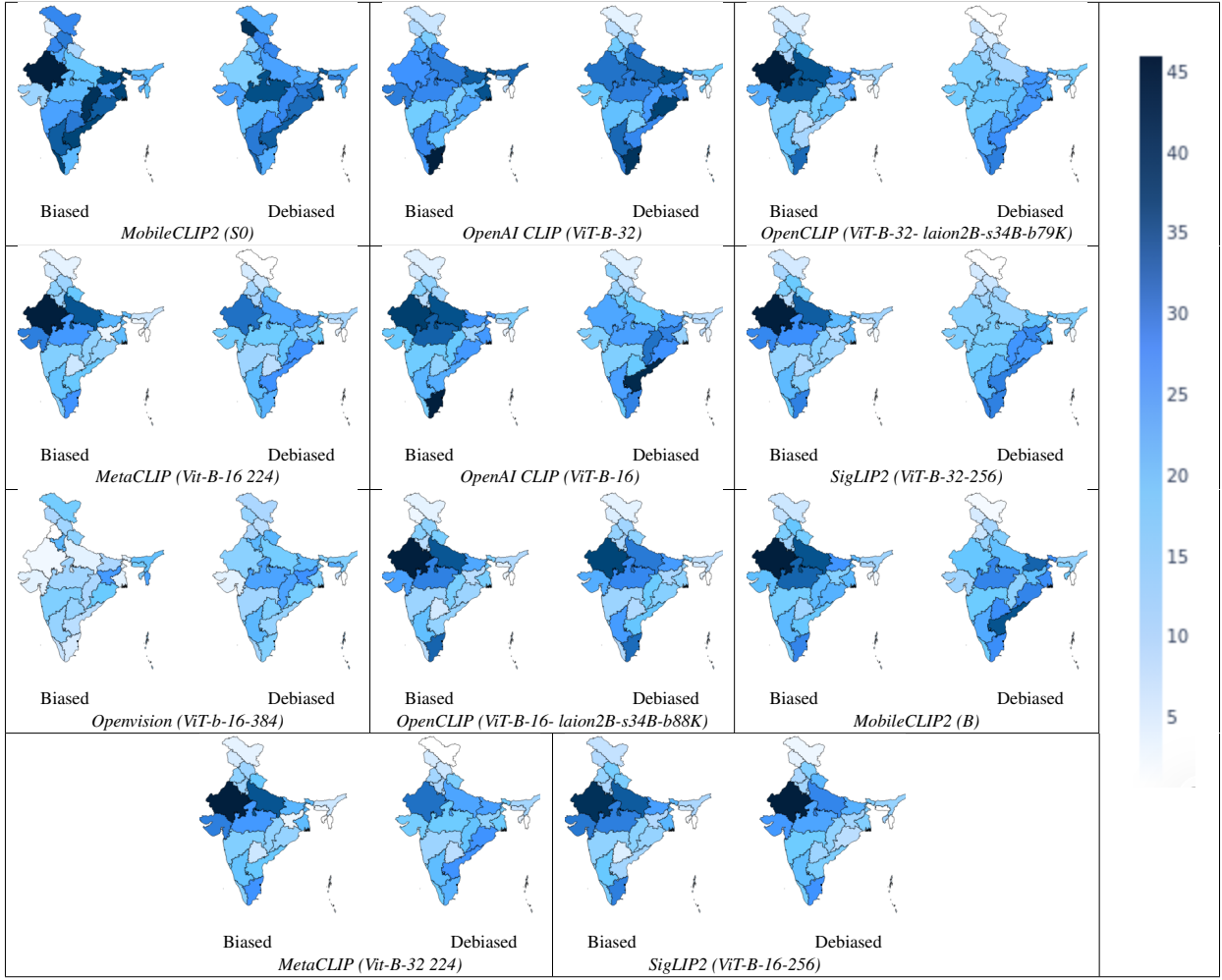

Figure 5: When prompted about "Indian person", certain states/regions such as Rajasthan, Uttar Pradesh, Tamil Nadu and Gujarat dominate the top-500 image retrievals. Post-INLP embeddings demonstrate improved geographical uniformity.

Table 3: Comparison of bias and debias statistics across vision-language models. Reported metrics include percentage reduction ($\Delta$) in dispersion (standard deviation $\sigma$) of average similarity score for each state before and after debiasing and percentage reduction in JSD values with respect to a uniform reference distribution ($\downarrow$ represents reduction for both dispersion and JSD values).

| Model | Prompts | | | | | |
| --- | --- | --- | --- | --- | --- | --- |
|  | Indian person | | Indian male | | Indian female | |
|  | $\Delta_\sigma \downarrow$ (in %) | JSD $\downarrow$ (in %) | $\Delta_\sigma \downarrow$ (in %) | JSD $\downarrow$ (in %) | $\Delta_\sigma \downarrow$ (in %) | JSD $\downarrow$ (in %) |
| MobileCLIP2 (S0) | 3.27 | 90.151 | 96.73 | 90.151 | 3.27 | 47.41 |
| OpenAI CLIP (ViT-B-32) | 13.149 | 77.959 | 86.851 | 77.959 | 13.149 | 353.704 |
| OpenCLIP (ViT-B-32-laion2B-s34B-b79K) | 22.24 | 74.108 | 77.76 | 74.108 | 22.24 | 286.217 |
| MetaCLIP (ViT-B-16-224) | 29.536 | 79.637 | 70.464 | 79.637 | 29.536 | 391.088 |
| OpenAI CLIP (ViT-B-16) | 20.331 | 81.371 | 79.669 | 81.371 | 20.331 | 436.804 |
| SigLIP2 (ViT-B-32-256) | 18.732 | 76.482 | 81.268 | 76.482 | 18.732 | 325.212 |
| OpenVision (ViT-B-16-384) | 8.171 | 8.639 | 99.76 | 8.639 | 0.24 | 9.456 |
| OpenCLIP (ViT-B-16-laion2B-s34B-b88K) | 20.569 | 66.516 | 79.431 | 66.516 | 20.569 | 198.054 |
| MobileCLIP (B) | 24.766 | 68.657 | 75.234 | 68.657 | 24.766 | 219.654 |
| MetaCLIP (ViT-B-32-224) | 23.221 | 76.312 | 76.749 | 76.312 | 23.221 | 322.151 |
| SigLIP2 (ViT-B-16-256) | 17.079 | 71.648 | 82.921 | 71.648 | 17.079 | 252.713 |

Table 4: Evaluation of generalization ability of debiased models on external classification benchmark datasets.

| Model | Original/ Debiased | Benchmark datasets | | | | | | | | |
|---|---|---|---|---|---|---|---|---|---|---|
| | | fer2013 | image net-a | image net-o | image net-r | image net1k | image net_sketch | image net-v2 | caltech 101 | cifar 10 | cifar 100 |
| MetaCLIP (ViT-B-16-224) | Original | 0.264 | 0.277 | 0.460 | 0.754 | 0.654 | 0.546 | 0.569 | 0.832 | 0.906 | 0.686 |
| | Debiased | 0.263 | 0.287 | 0.469 | 0.747 | 0.654 | 0.533 | 0.576 | 0.813 | 0.909 | 0.692 |
| OpenAI CLIP (ViT-B-16) | Original | 0.463 | 0.500 | 0.423 | 0.777 | 0.683 | 0.483 | 0.619 | 0.821 | 0.908 | 0.669 |
| | Debiased | 0.485 | 0.501 | 0.424 | 0.776 | 0.681 | 0.482 | 0.617 | 0.821 | 0.908 | 0.669 |
| OpenAI CLIP (ViT-B-32) | Original | 0.412 | 0.316 | 0.477 | 0.693 | 0.633 | 0.423 | 0.559 | 0.816 | 0.898 | 0.643 |
| | Debiased | 0.420 | 0.318 | 0.476 | 0.692 | 0.631 | 0.420 | 0.560 | 0.814 | 0.898 | 0.640 |
| OpenCLIP (ViT-B-16-laion2B-s34B-b88K) | Original | 0.520 | 0.373 | 0.465 | 0.817 | 0.706 | 0.575 | 0.629 | 0.838 | 0.944 | 0.761 |
| | Debiased | 0.517 | 0.375 | 0.459 | 0.816 | 0.706 | 0.574 | 0.628 | 0.838 | 0.944 | 0.760 |
| OpenCLIP (ViT-B-32-laion2B-s34B-b79K) | Original | 0.467 | 0.259 | 0.497 | 0.775 | 0.668 | 0.552 | 0.592 | 0.837 | 0.930 | 0.744 |
| | Debiased | 0.464 | 0.259 | 0.495 | 0.777 | 0.668 | 0.551 | 0.593 | 0.836 | 0.930 | 0.744 |
| OpenVision (ViT-B-16-384) | Original | 0.089 | 0.702 | 0.375 | 0.909 | 0.762 | 0.657 | 0.700 | 0.792 | 0.930 | 0.872 |
| | Debiased | 0.088 | 0.700 | 0.375 | 0.909 | 0.762 | 0.659 | 0.700 | 0.783 | 0.929 | 0.873 |
| SigLIP2 (ViT-B-16-256) | Original | 0.516 | 0.470 | 0.378 | 0.904 | 0.763 | 0.679 | 0.693 | 0.842 | 0.915 | 0.711 |
| | Debiased | 0.507 | 0.470 | 0.379 | 0.904 | 0.764 | 0.679 | 0.694 | 0.843 | 0.913 | 0.711 |

## 5 Results

### 5.1 Biased state representation

We first analyzed the top-500 retrievals from each VLM to quantify which regions dominate the models' perception of Indianness. The OpenCLIP (B16/B32), SigLIP (B16/B32), and MetaCLIP (B16/B32) families collectively retrieved more than 220-250 images ($\approx$ 45-50 %) from just five states i.e. Rajasthan, Uttar Pradesh, Tamil Nadu, Puducherry, and Gujarat, while the remaining 30+ states and Union Territories shared the residual half of the retrievals. Rajasthan alone appeared in 8–10 % of all top-ranked results per model ($\approx$ 40–50 images out of 500), followed by Uttar Pradesh and TamilNadu at around 7–9 % each. Figure 5 shows the state-wise distribution of top-500 image retrievals for different VLMs.

Conversely, the North-Eastern and peripheral regions such as Nagaland, Mizoram, Sikkim, Tripura, Arunachal Pradesh, and the island UT of Lakshadweep appeared only 1–3 times per model ($< 0.5$ %), often missing entirely in some variants. Even higher-capacity architectures such as OpenVision-Large failed to retrieve faces from these regions which reinforces the dominance of a narrow set of North-Western and Southern states. The lightweight models (such as MobileCLIP2-S0) exhibited some instability, retrieving clusters dominated by Arunachal Pradesh, Rajasthan, and Chhattisgarh. The pre-debiasing audit demonstrated that the concept of "Indianness" in current VLMs is geographically collapsed into a North-West–South corridor, over-representing culturally visible states like Rajasthan and Tamil Nadu while marginalizing the North-East, Himalayan, and island regions. The results of other prompts related to male and female gender were also similar (see Table 3 and refer *Supplementary Material*).

### 5.2 Debiasing

After applying the debiasing pipeline, the top-500 retrieval distributions became significantly more uniform across states. The dominance of a few regions evident before debiasing was reduced which suggests the directional removal of bias components in the CLIP embedding space effectively neutralized over-represented regions. Across all VLMs, the state-level dispersion in cosine similarity scores (Table 3) dropped i.e. the standard deviation of average per-state similarity decreased and the normalized JSD with respect to a uniform distribution also fell.

Quantitatively, the post-debiased retrievals redistributed samples much more evenly: previously dominant states such as Rajasthan, Uttar Pradesh, and Tamil Nadu each now accounted for 5–7 % of top-500 images ($\approx$ 25–35 per model), while under-represented regions including Nagaland, Mizoram, Tripura, and Sikkim—rose from $< 0.5$ % occurrence to 2-3 % and appearing consistently across most models. Notably, the North-Eastern and central states (e.g., Chhattisgarh, Jharkhand, Bihar) became more visible among top-retrieved results and reflect an expanded notion of Indianness within the embedding space.

### 5.3 Generalization of debiased embeddings

A major criticism of debiasing methods is that though they reduce unfair correlations, they often distort the embedding space which leads to degrading performance on unrelated downstream tasks. To test it, we evaluated both original and debiased VLMs on a comprehensive suite of 35 public classification benchmark datasets [2] (e.g., Cars, Country211, FER2013, Aircraft, ImageNet, CIFAR etc.). Across the datasets, the IndicFairFace-debiased embeddings preserved near-identical generalization ability. For instance, MetaCLIP-base maintained an average accuracy drop of only 0.13% across all benchmarks, while OpenCLIP (B16/B32) and OpenVision variants showed negligible deviation ($< 1\%$) from their original scores. At a few instances, the debiased models even showed slight positive gains. The consistency across 35 benchmarks indicates that the bias subspace identified by our debiasing approach is largely orthogonal to the primary semantic axes. A sample of results for few models and benchmark datasets is shown in Table 4.

Table 5: Accuracy of Vision-Language Models (VLMs) in the Job Application Bias Simulation across At-Risk Indian States, before and after debiasing.

| Model | State<br>Accuracy | Andaman and Nicobar Islands | Arunachal pradesh | Assam | Jammu and kashmir | Ladakh | Manipur | Meghalaya | Mizoram | Nagaland | Sikkim | Tripura |
|---|---|---|---|---|---|---|---|---|---|---|---|---|
| MobileCLIP2 (S0) | Base Acc | 0.000 | 0.000 | 0.000 | 0.000 | 0.000 | 0.000 | 0.000 | 0.000 | 0.000 | 0.000 | 0.000 |
|  | Debiased Acc | 0.772 | 0.814 | 0.629 | 0.672 | 0.689 | 0.765 | 0.802 | 0.773 | 0.714 | 0.838 | 0.707 |
| OpenAI CLIP (ViT-B-32) | Base Acc | 0.474 | 0.220 | 0.871 | 0.697 | 0.546 | 0.409 | 0.569 | 0.235 | 0.235 | 0.521 | 0.578 |
|  | Debiased Acc | 1.000 | 1.000 | 1.000 | 1.000 | 1.000 | 1.000 | 1.000 | 1.000 | 1.000 | 1.000 | 1.000 |
| OpenCLIP (ViT-B-32) laion2B-s34B-b79K) | Base Acc | 0.667 | 0.525 | 0.948 | 0.832 | 0.689 | 0.600 | 0.767 | 0.546 | 0.496 | 0.880 | 0.716 |
|  | Debiased Acc | 1.000 | 1.000 | 0.991 | 0.983 | 1.000 | 0.991 | 1.000 | 1.000 | 0.983 | 1.000 | 0.940 |
| MetaCLIP (Vit-B-16_224) | Base Acc | 0.439 | 0.237 | 0.836 | 0.613 | 0.429 | 0.391 | 0.560 | 0.235 | 0.328 | 0.419 | 0.560 |
|  | Debiased Acc | 1.000 | 1.000 | 1.000 | 1.000 | 1.000 | 1.000 | 1.000 | 1.000 | 1.000 | 1.000 | 1.000 |
| OpenAI CLIP (ViT-B-16) | Base Acc | 0.684 | 0.441 | 0.940 | 0.849 | 0.739 | 0.504 | 0.750 | 0.328 | 0.529 | 0.846 | 0.672 |
|  | Debiased Acc | 1.000 | 0.983 | 1.000 | 1.000 | 1.000 | 1.000 | 0.991 | 0.958 | 0.992 | 1.000 | 0.991 |
| SigLIP2 (ViT-B-32-256) | Base Acc | 0.719 | 0.847 | 0.957 | 0.916 | 0.664 | 0.835 | 0.853 | 0.773 | 0.555 | 0.855 | 0.853 |
|  | Debiased Acc | 1.000 | 1.000 | 1.000 | 1.000 | 1.000 | 1.000 | 1.000 | 1.000 | 1.000 | 1.000 | 1.000 |
| Openvision (ViT-b-16-384) | Base Acc | 0.000 | 0.000 | 0.000 | 0.008 | 0.000 | 0.035 | 0.052 | 0.008 | 0.000 | 0.009 | 0.000 |
|  | Debiased Acc | 0.997 | 1.000 | 0.988 | 0.995 | 1.000 | 0.985 | 0.992 | 1.000 | 0.998 | 0.989 | 0.994 |
| OpenCLIP (ViT-B-16) laion2B-s34B-b88K) | Base Acc | 0.561 | 0.492 | 0.931 | 0.849 | 0.739 | 0.652 | 0.810 | 0.563 | 0.454 | 0.880 | 0.776 |
|  | Debiased Acc | 1.000 | 1.000 | 1.000 | 1.000 | 1.000 | 1.000 | 1.000 | 1.000 | 1.000 | 1.000 | 1.000 |
| MobileCLIP2 (B) | Base Acc | 0.825 | 0.932 | 0.948 | 0.874 | 0.933 | 0.861 | 0.802 | 0.748 | 0.874 | 0.778 | 0.776 |
|  | Debiased Acc | 0.965 | 1.000 | 1.000 | 0.958 | 1.000 | 1.000 | 0.974 | 0.966 | 1.000 | 1.000 | 1.000 |
| MetaCLIP (Vit-B-32_224) | Base Acc | 0.368 | 0.059 | 0.586 | 0.361 | 0.202 | 0.278 | 0.457 | 0.151 | 0.092 | 0.231 | 0.517 |
|  | Debiased Acc | 1.000 | 1.000 | 1.000 | 0.992 | 1.000 | 0.991 | 1.000 | 1.000 | 1.000 | 0.991 | 1.000 |
| SigLIP2 (ViT-B-16-256) | Base Acc | 0.684 | 0.746 | 0.940 | 0.874 | 0.479 | 0.670 | 0.836 | 0.681 | 0.454 | 0.829 | 0.750 |
|  | Debiased Acc | 1.000 | 1.000 | 1.000 | 1.000 | 1.000 | 1.000 | 1.000 | 1.000 | 1.000 | 1.000 | 1.000 |

### 5.4 Downstream Task

To further validate the efficacy of our debiasing framework within the Indian context, we designed a zero-shot classification experiment that emulates a recruiter's decision-making process when distinguishing between Indian and non-Indian candidates. The task mirrors practical deployment scenarios such as identity verification, automated recruitment screening, or personalized recommendation systems where equitable geographic representation is essential. For the experiment, we specifically targeted the 11 "at-risk" Indian states that previously exhibited low "Indianness" similarity scores and higher misclassification tendencies (results indicated that VLMs strongly associated prominent states such as Gujrat, Uttar Pradesh etc. with "Indianness" even before debiasing). For each state, we curated representative facial samples and presented them to VLMs with two natural-language prompts: "Indian candidate" and "foreign candidate." The model's higher similarity between the prompts served as an indicator of whether it associates each state's visual identity with the broader semantic category of "Indian." The results are presented in Table 5.

## 6 Conclusion

Our work introduced IndicFairFace, the first geographically balanced face dataset designed to audit and mitigate intra-national bias in VLMs for the Indian context. Through comprehensive evaluation across multiple CLIP-based architectures, we demonstrated that existing VLMs encode a geographically skewed notion of Indianness, disproportionately favoring a narrow set of North-Western and Southern states while under-representing the North-Eastern and island regions. The adopted debiasing framework reduced inter-state variance and improved fairness metrics without compromising model utility. The debiased embeddings yielded better uniform retrievals and enhanced performance in bias-sensitive simulations such as the job application task, while maintaining strong generalization across external downstream benchmarks.

The proposed framework provided a foundation for region-aware fairness auditing and encourages future research on intersectional and cultural bias mitigation across multimodal AI systems.

# 7 Data Collection via Wikimedia API

To construct the *IndicFairFace* dataset, we developed a custom data extraction pipeline utilizing the Wikidata Query Service (WDQS). The pipeline was implemented in Python using the `SPARQLWrapper` library to interface with the public SPARQL endpoint.

## 7.1 SPARQL Query Construction

For each of the 28 States and 8 Union Territories of India, we constructed specific SPARQL queries to retrieve individuals associated with that region. To ensure balanced demographic representation at the source, we executed separate queries for binary gender categories. The query logic filtered entities based on the following primary predicates:

- **Instance of Human** ($P31$): Restricted to entities classified as humans ($Q5$).
- **Gender** ($P21$): Filtered explicitly for Male ($Q6581097$) or Female ($Q6581072$) identifiers to maintain gender parity in the raw collection.
- **Image Availability** ($P18$): Required the presence of an associated image property to ensure visual data availability.

## 7.2 Geographic Association Logic

A core challenge in curating a state-balanced dataset is identifying individuals who strongly represent a specific region. To address this, we employed a robust union of filtering conditions that link an individual to a state ($S_{id}$) through various properties. An individual was included if they satisfied any of the following associations:

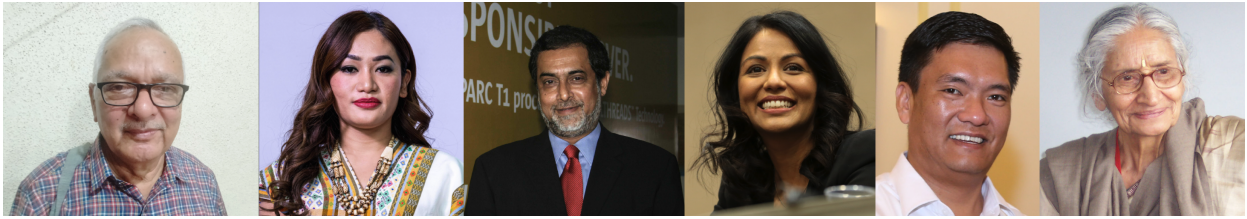

Figure 6: Sample of raw images before pre-processing

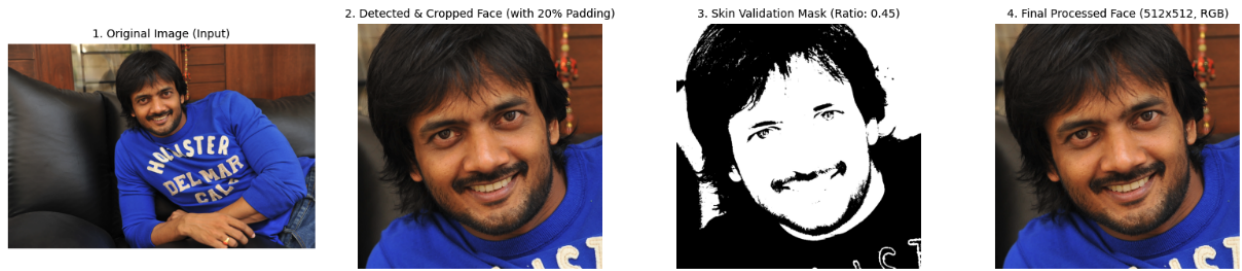

Figure 7: Steps in the preprocessing and augmentation Pipeline

1. **Direct Location Association:** The individual's Place of Birth ($P19$), Place of Residence ($P551$), or Work Location ($P937$) is located in ($P131*$) the target state. The transitive property path ($*$) allows for matching locations nested deeply within the state's administrative hierarchy (e.g., a specific village within a district of the state).

2. **Educational Association:** The individual was educated at ($P69$) an institution that is geographically located within the target state.

3. **Political Constituency:** The individual held a position ($P39$) representing a constituency ($P768$) located within the target state.

The resulting SPARQL query template for a given state $S$ and gender $G$ is formally defined as:

```
SELECT DISTINCT ?person ?personLabel
?image WHERE {
  ?person wdt:P31 wd:Q5;
          wdt:P21 wd:G;
          wdt:P18 ?image.
  {
    ?person wdt:P19|wdt:P551|wdt:P937
    ?loc.
    ?loc wdt:P131* wd:S.
  }
  UNION
  {
    ?person wdt:P69 ?inst.
    ?inst wdt:P131* wd:S.
  }
  UNION
  {
    ?person wdt:P39 ?pos.
    ?pos wdt:P768 ?const.
    ?const wdt:P131* wd:S.
  }
}
```

Table 6: Distribution of images in the Imbalanced dataset

| State | Male (Wikimedia) | Female (Wikimedia) | Male (SerAPI) | Female (SerAPI) |
|---|---|---|---|---|
| **Andaman and Nicobar Islands** | 1 | 0 | 33 | 52 |
| **Andhra Pradesh** | 100 | 97 | 70 | 84 |
| **Arunachal Pradesh** | 6 | 3 | 71 | 71 |
| **Assam** | 66 | 80 | 60 | 76 |
| **Bihar** | 57 | 62 | 66 | 79 |
| **Chandigarh** | 40 | 50 | 46 | 51 |
| **Chhattisgarh** | 11 | 14 | 51 | 72 |
| **Dadra and Nagar Haveli and Daman and Diu** | 1 | 1 | 23 | 12 |
| **Delhi** | 290 | 305 | 66 | 64 |
| **Goa** | 22 | 24 | 58 | 69 |
| **Gujarat** | 105 | 133 | 51 | 80 |
| **Haryana** | 73 | 81 | 60 | 60 |
| **Himachal Pradesh** | 30 | 35 | 74 | 77 |
| **Jammu and Kashmir** | 30 | 38 | 73 | 73 |
| **Jharkhand** | 34 | 41 | 62 | 59 |
| **Karnataka** | 273 | 309 | 66 | 70 |
| **Kerala** | 357 | 363 | 70 | 80 |
| **Ladakh** | 1 | 3 | 70 | 83 |
| **Lakshwadeep** | 1 | 2 | 37 | 38 |
| **Madhya Pradesh** | 73 | 84 | 59 | 71 |
| **Maharashtra** | 1109 | 1253 | 62 | 75 |
| **Manipur** | 21 | 17 | 65 | 66 |
| **Meghalaya** | 12 | 16 | 68 | 71 |
| **Mizoram** | 6 | 6 | 66 | 64 |
| **Nagaland** | 5 | 5 | 79 | 69 |
| **Odisha** | 89 | 101 | 73 | 54 |
| **Puducherry** | 6 | 11 | 68 | 75 |
| **Punjab** | 79 | 88 | 63 | 62 |
| **Rajasthan** | 66 | 71 | 62 | 77 |
| **Sikkim** | 1 | 1 | 57 | 62 |
| **Tamil Nadu** | 340 | 336 | 79 | 85 |
| **Telangana** | 146 | 157 | 53 | 58 |
| **Tripura** | 5 | 8 | 60 | 68 |
| **Uttar Pradesh** | 226 | 241 | 77 | 64 |
| **Uttarakhand** | 58 | 62 | 70 | 68 |
| **West Bengal** | 378 | 425 | 81 | 75 |

### 7.3 Execution and Aggregation

We iterated the whole query process over the set of Wikidata Q-IDs corresponding to all 36 administrative units (e.g., Uttar Pradesh: `Q1498`, Assam: `Q1164`). The raw results were retrieved in JSON format, transformed into Pandas DataFrames, and aggregated to form the initial candidate pool for the dataset. The distribution of the images extracted using these queries can be seen in Table 6. Examples of images extracted can be seen in Figure 6.

## 8 Preprocessing Pipeline

To ensure the dataset comprises realistic and geographically relevant facial images, we implemented a rigorous cleaning and standardization pipeline. As outlined in our methodology, raw images sourced from Wikimedia Commons and web scraping were processed using a multi-stage system built on OpenCV. The pipeline consists of three primary phases: candidate validation, face detection, and geometric standardization.

### 8.1 Quality and Content Validation

Before face detection, images underwent heuristic filtering to discard samples unsuitable for high-quality training data. This process was implemented using the following criteria:

- **Colour Consistency (Grayscale Rejection):** To maintain consistency in skin-tone representation, we excluded grayscale or black-and-white portraits. This was achieved by analyzing the inter-channel variance. For an image with Blue ($B$), Green ($G$), and Red ($R$) channels, we computed the mean absolute differences:

$$\delta_{color} = \frac{1}{3}\left(\overline{|B-G|} + \overline{|G-R|} + \overline{|R-B|}\right) \qquad (1)$$

Images where $\delta_{color}$ fell below a threshold of 5 were classified as monochromatic and rejected.

- **Texture Analysis (Synthetic Filtering):** To filter out cartoon-like, blurred, or synthetic faces, we utilized the variance of the Laplacian operator as a measure of focus and texture detail. Images exhibiting low Laplacian variance (indicating a lack of high-frequency edge information typical of real photographs) were discarded.
- **Skin Pixel Validation:** We applied HSV (Hue, Saturation, Value) skin-tone masking to ensure the candidate region contained a realistic ratio of skin pixels. A pixel-wise mask was generated within the range $H \in [0, 25]$, $S \in [40, 255]$, and $V \in [60, 255]$. Images were retained only if the ratio of non-zero mask pixels to total pixels exceeded a threshold of 0.15.

### 8.2 Face Detection

For accurate localization, we employed OpenCV's Deep Neural Network (DNN) module utilizing a ResNet-10 Single Shot Detector (SSD) model pre-trained on the Caffe framework ($300 \times 300$ input). This approach was chosen for its robustness over varied poses and lighting conditions compared to traditional Haar cascades.

The detector forward-passes the image blob to generate bounding box predictions. We filtered detections with a confidence score $< 0.5$. In scenarios with multiple detections, the pipeline selected the face with the largest bounding box area ($w \times h$) to ensure the primary subject was captured.

### 8.3 Geometric Standardization

Upon validating a face, we performed a crop-and-resize operation to standardize the dataset structure:

1. **Contextual Padding:** To preserve facial contours and avoid tight crops that eliminate features like hair or chin structure, we applied a dynamic padding margin. A padding of $20\%$ of the maximum dimension ($\max(w, h)$) was added to all sides of the bounding box coordinates.
2. **Resizing:** The cropped regions were resized to a uniform resolution of $512 \times 512$ pixels. To minimize aliasing and preserve feature fidelity during downsampling or upsampling, we utilized Lanczos interpolation.
3. **Format:** All processed images were converted and saved in the RGB color space.

This pipeline ensures that the final *IndicFairFace* dataset contains only valid, realistic, and standardized facial images suitable for training and auditing Vision-Language Models. The preprocessing steps are visualized in Figure 7.

## 9 Prompt Robustness Analysis

To ensure that the identified geographical biases were not artifacts of specific lexical choices, we evaluated the Vision-Language Models (VLMs) using a diverse set of synonymous prompts. We categorized these prompts into three groups: Neutral, Male, and Female, to analyze bias across gender lines comprehensively. The complete list of prompt variations tested is as follows:

- **Neutral-Gender Prompts (General Indian Context):** "An Indian person", "A person from India", "An Indian individual", "A person belonging to India", "A portrait of an Indian person", "A photograph of a person from India", "An Indian citizen", "A person native to India", "A resident of India", and "An Indian face portrait".
- **Male-Specific Prompts:** "An Indian man", "A man from India", "An Indian male", "A portrait of an Indian man", "A photograph of a man from India", "An Indian gentleman", "A male person belonging to India", "An Indian adult man", "A man native to India", and "A male citizen of India".
- **Female-Specific Prompts:** "An Indian woman", "A woman from India", "An Indian female", "A portrait of an Indian woman", "A photograph of a woman from India", "An Indian lady", "A female person belonging to India", "An Indian adult woman", "A woman native to India", and "A female citizen of India".

Our empirical analysis revealed that the retrieval distribution of Indian states remained consistent across these variations within each category. For instance, prompts such as "An Indian person" and "An Indian citizen" yielded statistically similar geographical skew, consistently over-representing North-Western and Southern regions while under-representing North-Eastern states. Given this high degree of consistency, we report our primary findings using the representative prompts "An Indian person", "An Indian male", and "An Indian female" for brevity, as the observed geographical bias patterns were robust to these phrasing variations.

## 10 Debiasing

The framework applies a geometric debiasing algorithm based on **Iterative Nullspace Projection (INLP)** combined with controlled interpolation and similarity compensation.

### 10.1 Overview

The first component of our debiasing pipeline targets bias removal directly in the embedding space of pre-trained CLIP models using linear algebraic projection techniques. The key idea is to identify one or more *bias directions*, vectors along which embeddings of biased and unbiased samples differ, and to project all embeddings onto the subspace orthogonal to these directions, effectively nullifying the bias component while preserving useful information.

### 10.2 Identifying Bias Directions

Let $E_0 = \{e_{0,i}\}_{i=1}^{N}$ denote embeddings representing the unbiased concept (e.g., "An Indian person") and $E_1 = \{e_{1,j}\}_{j=1}^{M}$ represent biased counterparts (e.g., "A Chinese person", "A Pakistani person"). A logistic regression classifier is trained to distinguish between $E_0$ and $E_1$ by solving the following binary classification problem:

$$P(y = 1 \mid e) = \sigma(w^T e + b) \tag{2}$$

where $\sigma(\cdot)$ is the sigmoid activation function. The learned weight vector $w$ corresponds to the direction of maximal separation between biased and unbiased embeddings. This vector is normalized to obtain the unit bias direction:

$$\hat{w} = \frac{w}{\|w\|_2} \tag{3}$$

### 10.3 Projection into the Nullspace

To remove the component of any embedding $v$ that lies along the bias direction $\hat{w}$, we project it onto the subspace orthogonal to $\hat{w}$. The projection matrix is given by:

$$P = I - \hat{w}\hat{w}^T \tag{4}$$

and the debiased embedding is computed as:

$$v_{\text{debiased}} = Pv = (I - \hat{w}\hat{w}^T)v \tag{5}$$

This operation preserves all information orthogonal to the bias direction while discarding any component aligned with it.

### 10.4 Iterative Debiasing Process

Since a single linear classifier may capture only one dominant bias direction, we repeat the procedure iteratively. After each projection, the embeddings are updated as:

$$E^{(k)} = E^{(k-1)} P_{k-1} \tag{6}$$

A new classifier is trained on $E^{(k)}$ to identify subsequent bias directions $w_k$. The process continues until the classifier's accuracy drops to approximately $0.5$, indicating that it can no longer distinguish between biased and unbiased embeddings. The overall transformation after $K$ iterations is:

$$v^{(K)} = P_{K-1} P_{K-2} \cdots P_0 v^{(0)} \tag{7}$$

### 10.5 Controlled Application via Spherical Interpolation (SLERP)

Completely removing the bias component can occasionally distort semantic relationships in the embedding space. To balance bias mitigation and utility preservation, we apply the projection partially using *Spherical Linear Interpolation (SLERP)* between the original and debiased embeddings This approach is effective as SLERP provides a principled way to interpolate on the embedding hypersphere.

Let $v_{\text{orig}}$ be the normalized original embedding and $v_{\text{proj}}$ the normalized debiased embedding. The interpolation angle is computed as:

$$\theta = \arccos(v_{\text{orig}} \cdot v_{\text{proj}}) \tag{8}$$

The interpolated embedding at debias strength $\alpha \in [0, 1]$ is:

$$v_{\text{interp}} = \frac{\sin((1-\alpha)\theta)}{\sin(\theta)} v_{\text{orig}} + \frac{\sin(\alpha\theta)}{\sin(\theta)} v_{\text{proj}} \tag{9}$$

A smaller $\alpha$ retains more of the original vector, while a larger $\alpha$ yields a stronger debiasing effect. The resulting embedding is rescaled to match the original magnitude:

$$v_{\text{final}} = v_{\text{interp}} \times \|v_{\text{orig, unnorm}}\|_2 \tag{10}$$

### 10.6 Similarity Compensation

Although projection effectively removes bias, it can also decrease the similarity of embeddings to the target text concept. To restore this alignment, we compute a *compensation vector* $c$ aligned with the target text embedding $t$:

$$\Delta S = S_{\text{orig}} - S_{\text{debiased}} \tag{11}$$
$$c = \beta t, \quad \beta = 2\Delta S \tag{12}$$

and adjust the final debiased embedding as:

$$v_{\text{comp}} = v_{\text{debiased}} + c \tag{13}$$

This addition ensures that while bias is minimized, the embedding remains semantically consistent.

## 11 Metrics

To assess the effectiveness of our debiasing pipeline, we evaluate the models on two primary axes: **utility** (the model's ability to perform its core task) and **fairness** (the uniformity of its performance across different demographic groups). Our evaluation relies on the following metrics, which are based on the cosine similarity ($\text{sim}(v, t)$) between image ($v$) and text ($t$) embeddings. Cosine Simimlarity is defined as:

$$\text{sim}(v, t) = \frac{v \cdot t}{\|v\|_2 \|t\|_2} \tag{14}$$

For each geographic subgroup (Indian state), the mean and standard deviation of similarity scores are computed. A lower inter-state standard deviation indicates improved fairness and reduced representational disparity across the country.

- **Average Similarity per Category (Mean Group Similarity):** This metric measures the model's average alignment with a target concept for a specific subgroup. For each demographic category $C_i$ (e.g., an Indian state), we compute the mean similarity between all its image embeddings $\{v_j \in C_i\}$ and the target text embedding $t$ (e.g., "A photo of an Indian person").

$$S_{\text{avg}}(C_i) = \frac{1}{|C_i|} \sum_{v_j \in C_i} \text{sim}(v_j, t) \tag{15}$$

  A high $S_{\text{avg}}$ across all categories indicates good overall utility. For fairness, we then compute the **standard deviation of these means** across all $N$ categories. A lower standard deviation indicates that all groups are recognized with similar confidence, signifying reduced representational disparity. This method is analogous to measuring performance gaps (e.g., in True Positive Rate) across groups in classification tasks.

- **Top-K Similarity (Recall@K):** It evaluates the practical retrieval utility of the model. For a given target text prompt $t$, we rank all images in a large, diverse validation set by their cosine similarity to $t$. We then measure the recall of relevant images within the top $K = 500$ retrieved results. A high score ensures that the debiasing process has not compromised the model's core ability to identify and rank correct images for a query.

- **Jensen-Shannon (JS) Score:** While the standard deviation of means compares average scores, the Jensen-Shannon Divergence (JSD) compares the full distributions of similarity scores between groups. JSD is a symmetric, smoothed and bounded measure of the statistical distance between two probability distributions and provides a measure of distributional fairness. We compute the JSD between the similarity score distribution of a specific subgroup ($P_{C_i}$) and a reference distribution ($P_{\text{ref}}$), which could be a baseline or an aggregated set.

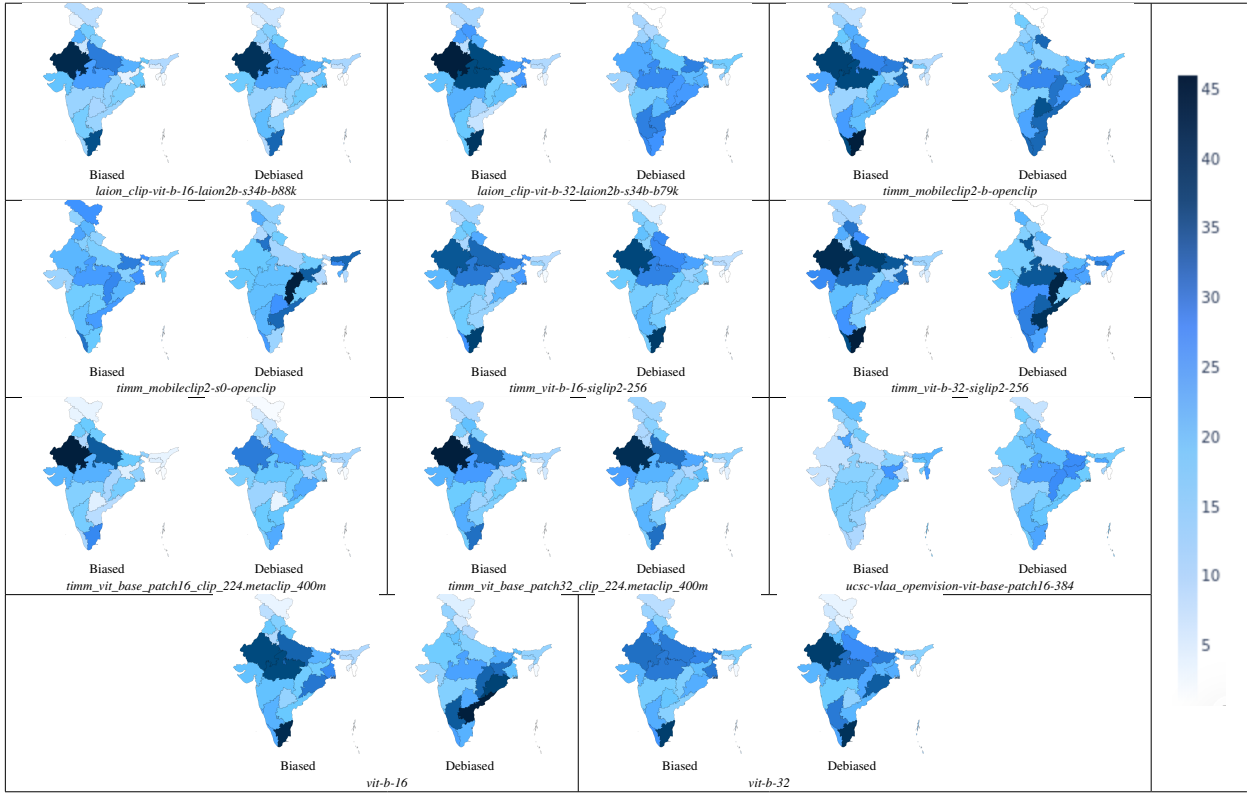

Figure 8: Distribution of images (biased and debiased) when prompted about "Indian male".

$$\text{JSD}(P_{C_i} \parallel P_{\text{ref}}) = H\left(\frac{P_{C_i} + P_{\text{ref}}}{2}\right) \\ - \frac{1}{2}\left[H(P_{C_i}) + H(P_{\text{ref}})\right] \quad (16)$$

where $H(P)$ is the Shannon entropy of a distribution. A JS score close to 0 indicates that the two distributions are nearly identical, demonstrating a high degree of fairness.

## 12 General Performance

For completeness, we report the full set of quantitative results in the Supplementary Material, where the final tables provide detailed state-wise distributions, debiasing metrics, and downstream evaluation scores. Figure 8 and 9 shows the state-wise distribution of top-500 images for different VLMs before and after debiasing. Table 7 presents the Top-1 accuracy and Table 8 reports the Top-5 accuracy of original and debiased VLMs on classification benchmarks. Table 9 summarizes the image-to-image retrieval performance and Table 10 provides the corresponding text-to-text retrieval accuracy of original and debiased VLMs. Table 11 reports the per-class recall values of original and debiased VLMs on classification benchmarks. All these measures indicate that debiasing had negligible impact on the general performance of VLMs.

## 13 Limitations

The authors acknowledge few limitations of the proposed work. The dataset primarily relies on publicly available images from Wikimedia Commons and open-license web repositories. While our approach ensures ethical sourcing and open licensing, the dataset may still exhibit a bias towards individuals who are more likely to have images in the public domain, such as politicians, celebrities and public figures. Moreover, for certain Indian states and Union Territories with insufficient Wikimedia coverage (e.g., Mizoram, Tripura, Ladakh, Andaman & Nicobar Islands), images were obtained through controlled web search methods and synthetic augmentations were introduced to achieve the desired balance. The reliance on web scraping and augmentation introduces a potential limitation in the representativeness of the raw and unaugmented data.

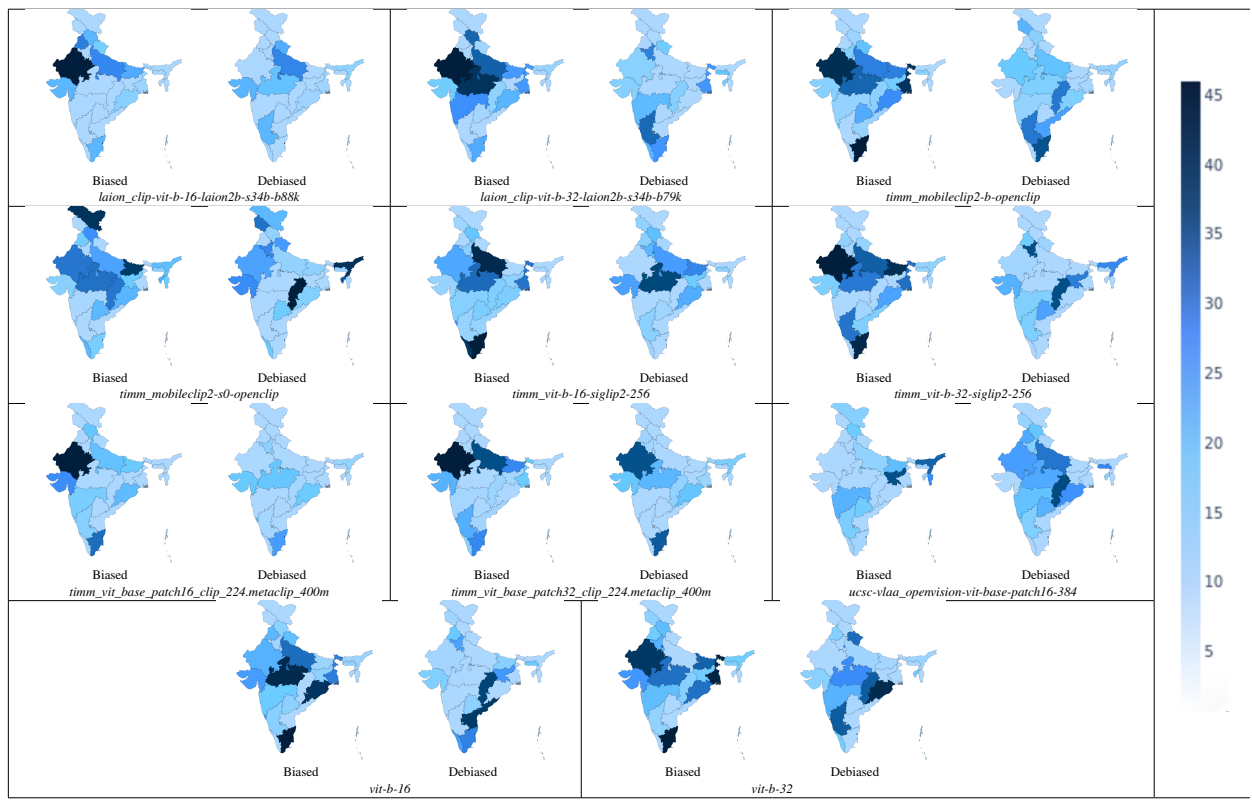

Figure 9: Distribution of images (biased and debiased) when prompted about "Indian female".

The current dataset is balanced across states/UTs using binary gender categories (Male and Female). It does not account for non-binary or other gender identities, which limits the scope of gender-based fairness studies to a binary perspective. Lastly, while the dataset provides balanced geographical representation, it does not explore finer-grained intra-state socio-cultural diversity such as caste, religion or language.

## 14 Ethics Statement

The IndicFairFace dataset was created with a commitment to ethical data practices, prioritising open licenses and geographical representation. All images were ethically sourced from Wikimedia Commons and other open-license repositories. Images from Wikimedia Commons were verified to use permissible open licenses such as CC0, CC-BY, or Public Domain. By sourcing images exclusively from openly licensed platforms, the authors ensure that individuals featured in the dataset had implicitly or explicitly agreed to public use and re-use of their images under the given licenses. The web-search pipeline additionally prioritized openly licensed content to maintain compliance with privacy standards.

Table 7: Dataset Benchmark Results: Classification Accuracy (Top 1)

| Dataset | MetaCLIP (ViT-B-16-224) — base | MetaCLIP (ViT-B-16-224) — finetuned | MetaCLIP (ViT-B-32-224) — base | MetaCLIP (ViT-B-32-224) — finetuned | MobileCLIP (B) — base | MobileCLIP (B) — finetuned | MobileCLIP2 (S0) — base | MobileCLIP2 (S0) — finetuned | OpenAI CLIP (ViT-B-16) — base | OpenAI CLIP (ViT-B-16) — finetuned | OpenAI CLIP (ViT-B-32) — base | OpenAI CLIP (ViT-B-32) — finetuned | OpenCLIP (ViT-B-16-laion2B-s34B-b88K) — base | OpenCLIP (ViT-B-16-laion2B-s34B-b88K) — finetuned | OpenCLIP (ViT-B-32-laion2B-s34B-b79K) — base | OpenCLIP (ViT-B-32-laion2B-s34B-b79K) — finetuned | OpenVision (ViT-B-16-384) — base | OpenVision (ViT-B-16-384) — finetuned | SigLIP2 (ViT-B-16-256) — base | SigLIP2 (ViT-B-16-256) — finetuned | SigLIP2 (ViT-B-32-256) — base | SigLIP2 (ViT-B-32-256) — finetuned |
|---|---|---|---|---|---|---|---|---|---|---|---|---|---|---|---|---|---|---|---|---|---|---|
| caltech101 | 0.805 | 0.805 | 0.832 | 0.813 | 0.800 | 0.800 | 0.780 | 0.780 | 0.821 | 0.821 | 0.816 | 0.814 | 0.838 | 0.838 | 0.837 | 0.836 | 0.792 | 0.783 | 0.842 | 0.843 | 0.881 | 0.881 |
| cars | 0.951 | 0.951 | 0.730 | 0.706 | 0.673 | 0.673 | 0.606 | 0.606 | 0.647 | 0.647 | 0.596 | 0.595 | 0.905 | 0.905 | 0.887 | 0.885 | 0.912 | 0.912 | 0.909 | 0.909 | 0.907 | 0.907 |
| cifar10 | 0.850 | 0.850 | 0.906 | 0.909 | 0.968 | 0.968 | 0.903 | 0.903 | 0.908 | 0.908 | 0.898 | 0.898 | 0.944 | 0.944 | 0.930 | 0.930 | 0.929 | 0.915 | 0.913 | 0.935 | 0.935 | 0.935 |
| cifar100 | 0.661 | 0.661 | 0.686 | 0.692 | 0.677 | 0.677 | 0.588 | 0.588 | 0.669 | 0.669 | 0.643 | 0.640 | 0.761 | 0.760 | 0.744 | 0.744 | 0.872 | 0.873 | 0.711 | 0.711 | 0.802 | 0.802 |
| clevr_closest_object_distance | 0.227 | 0.227 | 0.213 | 0.246 | 0.216 | 0.216 | 0.169 | 0.169 | 0.158 | 0.158 | 0.163 | 0.168 | 0.223 | 0.212 | 0.178 | 0.164 | 0.242 | 0.243 | 0.243 | 0.242 | 0.234 | 0.234 |
| clevr_count_all | 0.240 | 0.240 | 0.213 | 0.229 | 0.195 | 0.195 | 0.229 | 0.229 | 0.207 | 0.207 | 0.232 | 0.236 | 0.210 | 0.210 | 0.147 | 0.148 | 0.142 | 0.142 | 0.238 | 0.237 | 0.195 | 0.195 |
| country211 | 0.146 | 0.146 | 0.176 | 0.169 | 0.181 | 0.181 | 0.174 | 0.174 | 0.229 | 0.229 | 0.172 | 0.170 | 0.209 | 0.208 | 0.173 | 0.172 | 0.257 | 0.254 | 0.160 | 0.159 | 0.195 | 0.195 |
| diabetic_retinopathy | 0.552 | 0.552 | 0.151 | 0.342 | 0.150 | 0.150 | 0.385 | 0.385 | 0.035 | 0.035 | 0.454 | 0.370 | 0.672 | 0.672 | 0.735 | 0.735 | 0.198 | 0.094 | 0.601 | 0.542 | 0.614 | 0.614 |
| dmlab | 0.204 | 0.204 | 0.163 | 0.107 | 0.148 | 0.148 | 0.199 | 0.199 | 0.155 | 0.155 | 0.193 | 0.192 | 0.190 | 0.188 | 0.169 | 0.168 | 0.122 | 0.125 | 0.214 | 0.213 | 0.192 | 0.192 |
| dsprites_label_orientation | 0.031 | 0.031 | 0.025 | 0.021 | 0.023 | 0.023 | 0.026 | 0.026 | 0.023 | 0.023 | 0.025 | 0.025 | 0.028 | 0.029 | 0.035 | 0.035 | 0.025 | 0.023 | 0.034 | 0.033 | 0.026 | 0.026 |
| dsprites_label_x_position | 0.033 | 0.033 | 0.032 | 0.032 | 0.034 | 0.034 | 0.039 | 0.039 | 0.030 | 0.030 | 0.036 | 0.036 | 0.033 | 0.033 | 0.028 | 0.029 | 0.030 | 0.030 | 0.037 | 0.035 | 0.033 | 0.033 |
| dsprites_label_y_position | 0.031 | 0.031 | 0.032 | 0.032 | 0.030 | 0.030 | 0.031 | 0.031 | 0.031 | 0.031 | 0.032 | 0.032 | 0.035 | 0.034 | 0.032 | 0.032 | 0.032 | 0.032 | 0.032 | 0.032 | 0.033 | 0.033 |
| dtd | 0.706 | 0.706 | 0.530 | 0.528 | 0.507 | 0.507 | 0.437 | 0.437 | 0.449 | 0.449 | 0.444 | 0.441 | 0.565 | 0.566 | 0.563 | 0.558 | 0.636 | 0.635 | 0.647 | 0.652 | 0.545 | 0.545 |
| eurosat | 0.470 | 0.470 | 0.520 | 0.516 | 0.476 | 0.476 | 0.559 | 0.559 | 0.505 | 0.505 | 0.510 | 0.523 | 0.524 | 0.471 | 0.472 | 0.466 | 0.717 | 0.724 | 0.434 | 0.435 | 0.554 | 0.554 |
| fer2013 | 0.524 | 0.524 | 0.264 | 0.263 | 0.273 | 0.273 | 0.456 | 0.456 | 0.463 | 0.485 | 0.412 | 0.420 | 0.520 | 0.517 | 0.467 | 0.464 | 0.089 | 0.088 | 0.516 | 0.507 | 0.542 | 0.542 |
| fgvc_aircraft | 0.421 | 0.421 | 0.302 | 0.268 | 0.282 | 0.282 | 0.177 | 0.177 | 0.243 | 0.243 | 0.196 | 0.194 | 0.294 | 0.289 | 0.272 | 0.284 | 0.383 | 0.391 | 0.426 | 0.428 | 0.318 | 0.318 |
| flowers | 0.849 | 0.849 | 0.709 | 0.709 | 0.659 | 0.659 | 0.629 | 0.629 | 0.712 | 0.712 | 0.664 | 0.657 | 0.711 | 0.706 | 0.718 | 0.716 | 0.775 | 0.773 | 0.858 | 0.858 | 0.660 | 0.660 |
| gtsrb | 0.530 | 0.530 | 0.353 | 0.369 | 0.386 | 0.386 | 0.294 | 0.294 | 0.434 | 0.434 | 0.326 | 0.322 | 0.484 | 0.485 | 0.494 | 0.492 | 0.624 | 0.627 | 0.515 | 0.513 | 0.458 | 0.458 |
| imagenet-a | 0.504 | 0.504 | 0.277 | 0.287 | 0.289 | 0.289 | 0.298 | 0.298 | 0.500 | 0.501 | 0.316 | 0.318 | 0.373 | 0.375 | 0.259 | 0.259 | 0.702 | 0.700 | 0.470 | 0.470 | 0.369 | 0.369 |
| imagenet-o | 0.349 | 0.349 | 0.460 | 0.469 | 0.487 | 0.487 | 0.475 | 0.475 | 0.423 | 0.424 | 0.477 | 0.476 | 0.465 | 0.459 | 0.497 | 0.495 | 0.375 | 0.375 | 0.378 | 0.379 | 0.461 | 0.461 |
| imagenet-r | 0.988 | 0.988 | 0.754 | 0.747 | 0.724 | 0.724 | 0.742 | 0.742 | 0.777 | 0.776 | 0.693 | 0.692 | 0.817 | 0.816 | 0.775 | 0.777 | 0.909 | 0.909 | 0.904 | 0.904 | 0.822 | 0.822 |
| imagenet1k | 0.839 | 0.839 | 0.654 | 0.654 | 0.604 | 0.604 | 0.674 | 0.674 | 0.683 | 0.681 | 0.633 | 0.631 | 0.706 | 0.706 | 0.668 | 0.668 | 0.762 | 0.762 | 0.763 | 0.764 | 0.728 | 0.728 |
| imagenet_sketch | 0.646 | 0.646 | 0.546 | 0.533 | 0.592 | 0.592 | 0.457 | 0.457 | 0.483 | 0.482 | 0.423 | 0.420 | 0.575 | 0.574 | 0.551 | 0.551 | 0.657 | 0.659 | 0.679 | 0.679 | 0.542 | 0.542 |
| imagenetv2 | 0.667 | 0.667 | 0.569 | 0.576 | 0.540 | 0.540 | 0.543 | 0.543 | 0.619 | 0.617 | 0.559 | 0.560 | 0.629 | 0.628 | 0.592 | 0.593 | 0.700 | 0.700 | 0.693 | 0.694 | 0.579 | 0.579 |
| kitti_closest_vehicle_distance | 0.415 | 0.415 | 0.266 | 0.274 | 0.288 | 0.288 | 0.299 | 0.299 | 0.262 | 0.262 | 0.274 | 0.273 | 0.165 | 0.169 | 0.263 | 0.263 | 0.284 | 0.276 | 0.390 | 0.394 | 0.178 | 0.178 |
| mnist | 0.879 | 0.879 | 0.400 | 0.318 | 0.414 | 0.414 | 0.467 | 0.467 | 0.518 | 0.518 | 0.487 | 0.494 | 0.636 | 0.629 | 0.692 | 0.696 | 0.870 | 0.869 | 0.837 | 0.840 | 0.598 | 0.598 |
| pcam | 0.589 | 0.589 | 0.644 | 0.650 | 0.665 | 0.665 | 0.589 | 0.589 | 0.507 | 0.507 | 0.622 | 0.620 | 0.561 | 0.563 | 0.587 | 0.581 | 0.500 | 0.500 | 0.594 | 0.593 | 0.574 | 0.574 |
| pets | 0.867 | 0.867 | 0.887 | 0.880 | 0.823 | 0.823 | 0.804 | 0.804 | 0.890 | 0.890 | 0.872 | 0.871 | 0.913 | 0.913 | 0.905 | 0.907 | 0.925 | 0.925 | 0.943 | 0.943 | 0.858 | 0.858 |
| renderedsst2 | 0.573 | 0.573 | 0.535 | 0.522 | 0.496 | 0.496 | 0.564 | 0.564 | 0.606 | 0.606 | 0.586 | 0.563 | 0.578 | 0.581 | 0.568 | 0.569 | 0.611 | 0.612 | 0.518 | 0.522 | 0.546 | 0.546 |
| resisc45 | 0.599 | 0.599 | 0.604 | 0.602 | 0.660 | 0.660 | 0.488 | 0.488 | 0.583 | 0.583 | 0.537 | 0.536 | 0.627 | 0.626 | 0.606 | 0.607 | 0.637 | 0.627 | 0.639 | 0.640 | 0.603 | 0.603 |
| smallnorb_label_azimuth | 0.048 | 0.048 | 0.049 | 0.049 | 0.047 | 0.047 | 0.062 | 0.062 | 0.051 | 0.051 | 0.061 | 0.059 | 0.056 | 0.059 | 0.061 | 0.061 | 0.058 | 0.061 | 0.048 | 0.049 | 0.056 | 0.056 |
| smallnorb_label_elevation | 0.107 | 0.107 | 0.095 | 0.095 | 0.094 | 0.094 | 0.123 | 0.123 | 0.122 | 0.122 | 0.127 | 0.128 | 0.115 | 0.115 | 0.110 | 0.110 | 0.136 | 0.134 | 0.111 | 0.110 | 0.107 | 0.107 |
| stl10 | 0.970 | 0.970 | 0.952 | 0.964 | 0.973 | 0.973 | 1.000 | 1.000 | 0.983 | 0.983 | 0.971 | 0.978 | 0.977 | 0.964 | 0.963 | 0.941 | 0.926 | 0.980 | 0.888 | 0.888 | | |
| svhn | 0.531 | 0.531 | 0.289 | 0.281 | 0.312 | 0.312 | 0.118 | 0.118 | 0.312 | 0.312 | 0.134 | 0.120 | 0.507 | 0.497 | 0.403 | 0.397 | 0.626 | 0.625 | 0.535 | 0.521 | 0.517 | 0.517 |
| voc2007 | 0.697 | 0.697 | 0.748 | 0.732 | 0.764 | 0.764 | 0.742 | 0.742 | 0.783 | 0.779 | 0.764 | 0.761 | 0.818 | 0.817 | 0.817 | 0.817 | 0.852 | 0.849 | 0.689 | 0.687 | 0.819 | 0.819 |

Table 8: Dataset Benchmark Results: Classification Accuracy (Top 5)

| Dataset | MetaCLIP (ViT-B-16-224) — base | MetaCLIP (ViT-B-16-224) — finetuned | MetaCLIP (ViT-B-32-224) — base | MetaCLIP (ViT-B-32-224) — finetuned | MobileCLIP (B) — base | MobileCLIP (B) — finetuned | MobileCLIP2 (S0) — base | MobileCLIP2 (S0) — finetuned | OpenAI CLIP (ViT-B-16) — base | OpenAI CLIP (ViT-B-16) — finetuned | OpenAI CLIP (ViT-B-32) — base | OpenAI CLIP (ViT-B-32) — finetuned | OpenCLIP (ViT-B-16-laion2B-s34B-b88K) — base | OpenCLIP (ViT-B-16-laion2B-s34B-b88K) — finetuned | OpenCLIP (ViT-B-32-laion2B-s34B-b79K) — base | OpenCLIP (ViT-B-32-laion2B-s34B-b79K) — finetuned | OpenVision (ViT-B-16-384) — base | OpenVision (ViT-B-16-384) — finetuned | SigLIP2 (ViT-B-16-256) — base | SigLIP2 (ViT-B-16-256) — finetuned | SigLIP2 (ViT-B-32-256) — base | SigLIP2 (ViT-B-32-256) — finetuned |
|---|---|---|---|---|---|---|---|---|---|---|---|---|---|---|---|---|---|---|---|---|---|---|
| caltech101 | 0.894 | 0.805 | 0.959 | 0.950 | 1.000 | 0.900 | 0.870 | 0.870 | 0.956 | 0.956 | 0.947 | 0.946 | 0.953 | 0.953 | 0.951 | 0.949 | 0.931 | 0.931 | 0.947 | 0.947 | 0.929 | 0.836 |
| cars | 0.941 | 0.847 | 0.973 | 0.963 | 0.947 | 0.853 | 0.933 | 0.933 | 0.943 | 0.943 | 0.916 | 0.913 | 0.997 | 0.996 | 0.995 | 0.995 | 0.998 | 0.998 | 0.998 | 0.998 | 1.000 | 0.900 |
| cifar10 | 0.938 | 0.844 | 0.997 | 0.997 | 1.000 | 0.900 | 1.000 | 1.000 | 0.994 | 0.994 | 0.996 | 0.997 | 0.999 | 0.999 | 0.998 | 0.998 | 1.000 | 1.000 | 0.996 | 0.995 | 0.957 | 0.861 |
| cifar100 | 0.826 | 0.744 | 0.906 | 0.907 | 0.827 | 0.745 | 0.955 | 0.955 | 0.892 | 0.892 | 0.888 | 0.885 | 0.942 | 0.941 | 0.935 | 0.935 | 0.985 | 0.985 | 0.903 | 0.902 | 0.883 | 0.794 |
| clevr_closest_object_distance | 0.926 | 0.833 | 0.910 | 0.910 | 0.996 | 0.896 | 0.857 | 0.857 | 0.801 | 0.801 | 0.910 | 0.910 | 0.921 | 0.921 | 0.910 | 0.909 | 0.894 | 0.894 | 0.901 | 0.901 | 1.000 | 0.900 |
| clevr_count_all | 0.753 | 0.678 | 0.823 | 0.754 | 0.747 | 0.672 | 0.792 | 0.792 | 0.784 | 0.784 | 0.782 | 0.794 | 0.680 | 0.678 | 0.701 | 0.705 | 0.762 | 0.761 | 0.763 | 0.758 | 0.650 | 0.585 |
| country211 | 0.389 | 0.351 | 0.397 | 0.386 | 0.358 | 0.322 | 0.392 | 0.392 | 0.485 | 0.485 | 0.404 | 0.401 | 0.446 | 0.443 | 0.389 | 0.387 | 0.518 | 0.517 | 0.361 | 0.360 | 0.425 | 0.382 |
| diabetic_retinopathy | 1.000 | 0.900 | 1.000 | 1.000 | 1.000 | 0.900 | 1.000 | 1.000 | 1.000 | 1.000 | 1.000 | 1.000 | 1.000 | 1.000 | 1.000 | 1.000 | 1.000 | 1.000 | 1.000 | 1.000 | 0.971 | 0.873 |
| dmlab | 0.769 | 0.693 | 0.830 | 0.780 | 0.856 | 0.771 | 0.823 | 0.823 | 0.841 | 0.841 | 0.851 | 0.853 | 0.842 | 0.840 | 0.820 | 0.815 | 0.795 | 0.792 | 0.813 | 0.813 | 0.874 | 0.787 |
| dsprites_label_orientation | 0.149 | 0.134 | 0.129 | 0.126 | 0.125 | 0.112 | 0.111 | 0.111 | 0.109 | 0.109 | 0.120 | 0.119 | 0.124 | 0.124 | 0.142 | 0.140 | 0.131 | 0.131 | 0.142 | 0.144 | 0.120 | 0.108 |
| dsprites_label_x_position | 0.170 | 0.153 | 0.157 | 0.143 | 0.129 | 0.160 | 0.160 | 0.151 | 0.151 | 0.172 | 0.172 | 0.160 | 0.159 | 0.146 | 0.147 | 0.158 | 0.158 | 0.170 | 0.168 | 0.151 | 0.136 | |
| dsprites_label_y_position | 0.147 | 0.132 | 0.167 | 0.167 | 0.161 | 0.145 | 0.144 | 0.144 | 0.165 | 0.165 | 0.155 | 0.156 | 0.170 | 0.171 | 0.160 | 0.161 | 0.146 | 0.148 | 0.156 | 0.156 | 0.156 | 0.140 |
| dtd | 0.919 | 0.827 | 0.830 | 0.830 | 0.793 | 0.714 | 0.726 | 0.726 | 0.758 | 0.758 | 0.766 | 0.765 | 0.862 | 0.858 | 0.871 | 0.869 | 0.907 | 0.907 | 0.925 | 0.924 | 0.890 | 0.801 |
| eurosat | 0.895 | 0.805 | 0.920 | 0.918 | 0.970 | 0.873 | 0.916 | 0.916 | 0.882 | 0.882 | 0.924 | 0.925 | 0.891 | 0.891 | 0.928 | 0.929 | 0.994 | 0.994 | 0.831 | 0.839 | 0.935 | 0.842 |
| fer2013 | 0.943 | 0.848 | 0.911 | 0.914 | 1.000 | 0.900 | 0.888 | 0.888 | 0.956 | 0.957 | 0.948 | 0.940 | 0.937 | 0.937 | 0.962 | 0.961 | 0.994 | 0.994 | 0.954 | 0.952 | 1.000 | 0.900 |
| fgvc_aircraft | 0.743 | 0.668 | 0.688 | 0.622 | 0.740 | 0.666 | 0.454 | 0.454 | 0.604 | 0.604 | 0.497 | 0.500 | 0.657 | 0.661 | 0.638 | 0.639 | 0.814 | 0.814 | 0.776 | 0.773 | 0.665 | 0.599 |
| flowers | 0.916 | 0.825 | 0.868 | 0.869 | 0.924 | 0.832 | 0.853 | 0.853 | 0.860 | 0.860 | 0.857 | 0.875 | 0.884 | 0.884 | 0.875 | 0.876 | 0.898 | 0.899 | 0.955 | 0.955 | 0.800 | 0.720 |
| gtsrb | 0.768 | 0.691 | 0.725 | 0.740 | 0.783 | 0.705 | 0.693 | 0.693 | 0.710 | 0.710 | 0.720 | 0.719 | 0.758 | 0.762 | 0.743 | 0.741 | 0.852 | 0.853 | 0.800 | 0.796 | 0.763 | 0.686 |
| imagenet-a | 0.771 | 0.694 | 0.599 | 0.615 | 0.658 | 0.592 | 0.630 | 0.630 | 0.805 | 0.803 | 0.642 | 0.644 | 0.694 | 0.696 | 0.566 | 0.567 | 0.912 | 0.913 | 0.751 | 0.751 | 0.728 | 0.655 |
| imagenet-o | 0.646 | 0.582 | 0.761 | 0.767 | 0.705 | 0.635 | 0.842 | 0.842 | 0.710 | 0.713 | 0.782 | 0.781 | 0.767 | 0.768 | 0.815 | 0.805 | 0.682 | 0.686 | 0.705 | 0.707 | 0.801 | 0.721 |
| imagenet-r | 0.992 | 0.893 | 0.911 | 0.907 | 0.877 | 0.789 | 0.935 | 0.935 | 0.930 | 0.930 | 0.889 | 0.888 | 0.946 | 0.946 | 0.921 | 0.920 | 0.983 | 0.983 | 0.975 | 0.975 | 0.934 | 0.841 |
| imagenet1k | 1.000 | 0.900 | 0.891 | 0.892 | 0.964 | 0.867 | 0.924 | 0.924 | 0.919 | 0.917 | 0.888 | 0.887 | 0.919 | 0.919 | 0.902 | 0.901 | 0.951 | 0.951 | 0.945 | 0.945 | 0.882 | 0.793 |
| imagenet_sketch | 0.945 | 0.850 | 0.802 | 0.793 | 0.784 | 0.705 | 0.732 | 0.732 | 0.763 | 0.761 | 0.703 | 0.703 | 0.830 | 0.830 | 0.805 | 0.804 | 0.881 | 0.883 | 0.889 | 0.889 | 0.816 | 0.735 |
| imagenetv2 | 0.826 | 0.744 | 0.833 | 0.835 | 0.911 | 0.820 | 0.775 | 0.775 | 0.873 | 0.873 | 0.864 | 0.833 | 0.872 | 0.872 | 0.845 | 0.843 | 0.915 | 0.917 | 0.909 | 0.909 | 0.895 | 0.805 |
| mnist | 0.973 | 0.875 | 0.785 | 0.800 | 0.750 | 0.675 | 0.855 | 0.855 | 0.868 | 0.868 | 0.851 | 0.843 | 0.881 | 0.881 | 0.939 | 0.943 | 0.980 | 0.980 | 0.979 | 0.980 | 0.938 | 0.844 |
| pets | 1.000 | 0.900 | 0.997 | 0.997 | 0.993 | 0.893 | 1.000 | 1.000 | 0.994 | 0.994 | 0.993 | 0.992 | 0.998 | 0.998 | 0.996 | 0.996 | 0.998 | 0.998 | 0.998 | 0.998 | 0.953 | 0.858 |
| resisc45 | 0.916 | 0.824 | 0.909 | 0.910 | 0.882 | 0.794 | 0.849 | 0.849 | 0.913 | 0.913 | 0.868 | 0.866 | 0.916 | 0.916 | 0.917 | 0.917 | 0.911 | 0.907 | 0.922 | 0.921 | 0.925 | 0.833 |
| smallnorb_label_azimuth | 0.297 | 0.268 | 0.271 | 0.272 | 0.246 | 0.221 | 0.286 | 0.286 | 0.288 | 0.288 | 0.287 | 0.278 | 0.280 | 0.272 | 0.266 | 0.265 | 0.285 | 0.283 | 0.305 | 0.274 | | |
| smallnorb_label_elevation | 0.592 | 0.532 | 0.540 | 0.539 | 0.563 | 0.507 | 0.541 | 0.541 | 0.562 | 0.562 | 0.584 | 0.582 | 0.535 | 0.536 | 0.580 | 0.578 | 0.564 | 0.562 | 0.565 | 0.564 | 0.509 | 0.458 |
| stl10 | 1.000 | 0.900 | 0.999 | 1.000 | 0.891 | 1.000 | 1.000 | 1.000 | 1.000 | 1.000 | 1.000 | 1.000 | 1.000 | 0.999 | 0.999 | 1.000 | 1.000 | 1.000 | 1.000 | 0.987 | 0.888 | |
| svhn | 0.862 | 0.776 | 0.762 | 0.755 | 0.809 | 0.728 | 0.599 | 0.599 | 0.765 | 0.765 | 0.652 | 0.648 | 0.810 | 0.805 | 0.791 | 0.786 | 0.958 | 0.957 | 0.874 | 0.872 | 0.828 | 0.745 |
| voc2007 | 0.817 | 0.736 | 0.937 | 0.927 | 0.894 | 0.805 | 0.870 | 0.870 | 0.957 | 0.955 | 0.959 | 0.960 | 0.969 | 0.969 | 0.969 | 0.969 | 0.989 | 0.989 | 0.897 | 0.896 | 0.889 | 0.801 |
| stl10 | 0.970 | 0.970 | 0.952 | 0.964 | 0.973 | 0.973 | 1.000 | 1.000 | 0.983 | 0.983 | 0.971 | 0.971 | 0.978 | 0.977 | 0.964 | 0.963 | 0.941 | 0.926 | 0.980 | 0.980 | 0.888 | 0.888 |
| svhn | 0.531 | 0.531 | 0.289 | 0.281 | 0.312 | 0.312 | 0.118 | 0.118 | 0.312 | 0.312 | 0.134 | 0.120 | 0.507 | 0.497 | 0.403 | 0.397 | 0.626 | 0.625 | 0.535 | 0.521 | 0.517 | 0.517 |
| voc2007 | 0.697 | 0.697 | 0.748 | 0.732 | 0.764 | 0.764 | 0.742 | 0.742 | 0.783 | 0.779 | 0.764 | 0.761 | 0.818 | 0.817 | 0.817 | 0.817 | 0.852 | 0.849 | 0.689 | 0.687 | 0.819 | 0.819 |

Table 9: Dataset Benchmark Results: Image to Image Retreival

| Dataset | MetaCLIP (ViT-B-16-224) — base | MetaCLIP (ViT-B-16-224) — finetuned | MetaCLIP (ViT-B-32-224) — base | MetaCLIP (ViT-B-32-224) — finetuned | MobileCLIP (B) — base | MobileCLIP (B) — finetuned | MobileCLIP2 (S0) — base | MobileCLIP2 (S0) — finetuned | OpenAI CLIP (ViT-B-16) — base | OpenAI CLIP (ViT-B-16) — finetuned | OpenAI CLIP (ViT-B-32) — base | OpenAI CLIP (ViT-B-32) — finetuned | OpenCLIP (ViT-B-16-laion2B-s34B-b88K) — base | OpenCLIP (ViT-B-16-laion2B-s34B-b88K) — finetuned | OpenCLIP (ViT-B-32-laion2B-s34B-b79K) — base | OpenCLIP (ViT-B-32-laion2B-s34B-b79K) — finetuned | OpenVision (ViT-B-16-384) — base | OpenVision (ViT-B-16-384) — finetuned | SigLIP2 (ViT-B-16-256) — base | SigLIP2 (ViT-B-16-256) — finetuned | SigLIP2 (ViT-B-32-256) — base | SigLIP2 (ViT-B-32-256) — finetuned |
|---|---|---|---|---|---|---|---|---|---|---|---|---|---|---|---|---|---|---|---|---|---|---|
| flickr30k | 0.926 | 0.926 | 0.858 | 0.853 | 0.858 | 0.858 | 0.826 | 0.826 | 0.856 | 0.853 | 0.834 | 0.826 | 0.909 | 0.908 | 0.888 | 0.886 | 0.952 | 0.953 | 0.927 | 0.926 | 0.909 | 0.909 |
| flickr8k | 0.920 | 0.920 | 0.846 | 0.838 | 0.846 | 0.846 | 0.796 | 0.796 | 0.829 | 0.822 | 0.805 | 0.796 | 0.896 | 0.895 | 0.873 | 0.868 | 0.936 | 0.937 | 0.921 | 0.920 | 0.896 | 0.896 |
| mscoco_captions | 0.721 | 0.721 | 0.628 | 0.616 | 0.628 | 0.628 | 0.552 | 0.552 | 0.584 | 0.578 | 0.560 | 0.552 | 0.687 | 0.687 | 0.668 | 0.662 | 0.768 | 0.763 | 0.724 | 0.721 | 0.687 | 0.687 |

Table 10: Dataset Benchmark Results: Text to Text Retreival

| Dataset | MetaCLIP (ViT-B-16-224) — base | MetaCLIP (ViT-B-16-224) — finetuned | MetaCLIP (ViT-B-32-224) — base | MetaCLIP (ViT-B-32-224) — finetuned | MobileCLIP (B) — base | MobileCLIP (B) — finetuned | MobileCLIP2 (S0) — base | MobileCLIP2 (S0) — finetuned | OpenAI CLIP (ViT-B-16) — base | OpenAI CLIP (ViT-B-16) — finetuned | OpenAI CLIP (ViT-B-32) — base | OpenAI CLIP (ViT-B-32) — finetuned | OpenCLIP (ViT-B-16-laion2B-s34B-b88K) — base | OpenCLIP (ViT-B-16-laion2B-s34B-b88K) — finetuned | OpenCLIP (ViT-B-32-laion2B-s34B-b79K) — base | OpenCLIP (ViT-B-32-laion2B-s34B-b79K) — finetuned | OpenVision (ViT-B-16-384) — base | OpenVision (ViT-B-16-384) — finetuned | SigLIP2 (ViT-B-16-256) — base | SigLIP2 (ViT-B-16-256) — finetuned | SigLIP2 (ViT-B-32-256) — base | SigLIP2 (ViT-B-32-256) — finetuned |
|---|---|---|---|---|---|---|---|---|---|---|---|---|---|---|---|---|---|---|---|---|---|---|
| flickr30k | 0.980 | 0.980 | 0.938 | 0.936 | 0.938 | 0.938 | 0.943 | 0.943 | 0.962 | 0.961 | 0.946 | 0.943 | 0.980 | 0.980 | 0.966 | 0.973 | 0.998 | 0.999 | 0.981 | 0.980 | 0.980 | 0.980 |
| flickr8k | 0.967 | 0.967 | 0.919 | 0.919 | 0.919 | 0.919 | 0.911 | 0.911 | 0.914 | 0.910 | 0.914 | 0.911 | 0.969 | 0.967 | 0.950 | 0.952 | 0.985 | 0.984 | 0.965 | 0.967 | 0.969 | 0.969 |
| mscoco_captions | 0.854 | 0.854 | 0.772 | 0.765 | 0.772 | 0.772 | 0.746 | 0.746 | 0.767 | 0.765 | 0.749 | 0.746 | 0.829 | 0.828 | 0.809 | 0.815 | 0.873 | 0.879 | 0.854 | 0.854 | 0.829 | 0.829 |

Table 11: Dataset Benchmark Results: Per Class Recall

| Dataset | MetaCLIP (ViT-B-16-224) — base | MetaCLIP (ViT-B-16-224) — finetuned | MetaCLIP (ViT-B-32-224) — base | MetaCLIP (ViT-B-32-224) — finetuned | MobileCLIP (B) — base | MobileCLIP (B) — finetuned | MobileCLIP2 (S0) — base | MobileCLIP2 (S0) — finetuned | OpenAI CLIP (ViT-B-16) — base | OpenAI CLIP (ViT-B-16) — finetuned | OpenAI CLIP (ViT-B-32) — base | OpenAI CLIP (ViT-B-32) — finetuned | OpenCLIP (ViT-B-16-laion2B-s34B-b88K) — base | OpenCLIP (ViT-B-16-laion2B-s34B-b88K) — finetuned | OpenCLIP (ViT-B-32-laion2B-s34B-b79K) — base | OpenCLIP (ViT-B-32-laion2B-s34B-b79K) — finetuned | OpenVision (ViT-B-16-384) — base | OpenVision (ViT-B-16-384) — finetuned | SigLIP2 (ViT-B-16-256) — base | SigLIP2 (ViT-B-16-256) — finetuned | SigLIP2 (ViT-B-32-256) — base | SigLIP2 (ViT-B-32-256) — finetuned |
|---|---|---|---|---|---|---|---|---|---|---|---|---|---|---|---|---|---|---|---|---|---|---|
| caltech101 | 0.976 | 0.976 | 0.924 | 0.913 | 0.835 | 0.835 | 0.948 | 0.853 | 0.890 | 0.890 | 0.876 | 0.876 | 0.929 | 0.929 | 0.920 | 0.918 | 0.928 | 0.923 | 0.951 | 0.951 | 1.000 | 1.000 |
| cars | 0.833 | 0.833 | 0.728 | 0.706 | 0.750 | 0.750 | 0.653 | 0.588 | 0.647 | 0.647 | 0.598 | 0.597 | 0.905 | 0.906 | 0.888 | 0.887 | 0.912 | 0.911 | 0.910 | 0.910 | 0.862 | 0.862 |
| cifar10 | 0.877 | 0.877 | 0.906 | 0.909 | 0.826 | 0.826 | 0.867 | 0.780 | 0.908 | 0.908 | 0.899 | 0.898 | 0.944 | 0.944 | 0.930 | 0.931 | 0.931 | 0.929 | 0.915 | 0.913 | 0.976 | 0.976 |
| cifar100 | 0.684 | 0.684 | 0.687 | 0.692 | 0.745 | 0.745 | 0.581 | 0.523 | 0.669 | 0.669 | 0.642 | 0.640 | 0.761 | 0.760 | 0.744 | 0.743 | 0.872 | 0.873 | 0.711 | 0.711 | 0.757 | 0.757 |
| clevr_closest_object_distance | 0.198 | 0.198 | 0.169 | 0.170 | 0.182 | 0.182 | 0.150 | 0.135 | 0.168 | 0.168 | 0.163 | 0.165 | 0.163 | 0.162 | 0.135 | 0.130 | 0.224 | 0.226 | 0.191 | 0.191 | 0.171 | 0.171 |
| clevr_count_all | 0.246 | 0.246 | 0.214 | 0.228 | 0.220 | 0.220 | 0.217 | 0.196 | 0.207 | 0.207 | 0.228 | 0.232 | 0.205 | 0.207 | 0.144 | 0.143 | 0.143 | 0.143 | 0.237 | 0.236 | 0.189 | 0.189 |
| country211 | 0.159 | 0.159 | 0.176 | 0.169 | 0.191 | 0.191 | 0.153 | 0.138 | 0.229 | 0.229 | 0.172 | 0.170 | 0.209 | 0.208 | 0.174 | 0.173 | 0.257 | 0.254 | 0.160 | 0.159 | 0.188 | 0.188 |
| diabetic_retinopathy | 0.224 | 0.224 | 0.246 | 0.250 | 0.252 | 0.252 | 0.207 | 0.186 | 0.210 | 0.210 | 0.222 | 0.219 | 0.294 | 0.292 | 0.200 | 0.200 | 0.241 | 0.243 | 0.232 | 0.232 | 0.296 | 0.296 |
| dmlab | 0.156 | 0.156 | 0.142 | 0.126 | 0.150 | 0.150 | 0.150 | 0.135 | 0.168 | 0.168 | 0.163 | 0.162 | 0.166 | 0.165 | 0.164 | 0.165 | 0.151 | 0.151 | 0.165 | 0.165 | 0.167 | 0.167 |
| dsprites_label_orientation | 0.036 | 0.036 | 0.025 | 0.021 | 0.024 | 0.024 | 0.023 | 0.021 | 0.023 | 0.023 | 0.025 | 0.025 | 0.027 | 0.029 | 0.035 | 0.035 | 0.025 | 0.022 | 0.034 | 0.033 | 0.025 | 0.025 |
| dsprites_label_x_position | 0.032 | 0.032 | 0.032 | 0.032 | 0.030 | 0.030 | 0.033 | 0.030 | 0.030 | 0.030 | 0.035 | 0.036 | 0.033 | 0.033 | 0.028 | 0.029 | 0.031 | 0.031 | 0.035 | 0.035 | 0.030 | 0.030 |
| dsprites_label_y_position | 0.028 | 0.028 | 0.031 | 0.031 | 0.034 | 0.034 | 0.031 | 0.028 | 0.031 | 0.031 | 0.031 | 0.031 | 0.031 | 0.035 | 0.034 | 0.032 | 0.031 | 0.031 | 0.031 | 0.031 | 0.037 | 0.037 |
| dtd | 0.674 | 0.674 | 0.530 | 0.527 | 0.484 | 0.484 | 0.404 | 0.364 | 0.451 | 0.451 | 0.444 | 0.440 | 0.565 | 0.566 | 0.563 | 0.559 | 0.637 | 0.635 | 0.647 | 0.651 | 0.593 | 0.593 |
| eurosat | 0.421 | 0.421 | 0.515 | 0.512 | 0.564 | 0.564 | 0.472 | 0.425 | 0.555 | 0.555 | 0.491 | 0.494 | 0.540 | 0.541 | 0.484 | 0.479 | 0.703 | 0.712 | 0.445 | 0.446 | 0.496 | 0.496 |
| fer2013 | 0.449 | 0.449 | 0.291 | 0.299 | 0.278 | 0.278 | 0.373 | 0.335 | 0.419 | 0.428 | 0.362 | 0.369 | 0.466 | 0.465 | 0.423 | 0.428 | 0.219 | 0.219 | 0.445 | 0.441 | 0.453 | 0.453 |
| fgvc_aircraft | 0.399 | 0.399 | 0.304 | 0.270 | 0.300 | 0.300 | 0.206 | 0.186 | 0.243 | 0.243 | 0.195 | 0.195 | 0.293 | 0.287 | 0.273 | 0.283 | 0.382 | 0.390 | 0.426 | 0.429 | 0.320 | 0.320 |
| flowers | 0.810 | 0.810 | 0.680 | 0.682 | 0.647 | 0.647 | 0.597 | 0.538 | 0.692 | 0.692 | 0.666 | 0.662 | 0.712 | 0.704 | 0.705 | 0.706 | 0.757 | 0.756 | 0.861 | 0.861 | 0.668 | 0.668 |
| gtsrb | 0.499 | 0.499 | 0.338 | 0.351 | 0.318 | 0.318 | 0.286 | 0.257 | 0.365 | 0.365 | 0.305 | 0.304 | 0.461 | 0.466 | 0.437 | 0.436 | 0.559 | 0.562 | 0.492 | 0.492 | 0.432 | 0.432 |
| imagenet-a | 0.461 | 0.461 | 0.309 | 0.300 | 0.283 | 0.283 | 0.322 | 0.290 | 0.483 | 0.485 | 0.327 | 0.328 | 0.403 | 0.402 | 0.305 | 0.301 | 0.691 | 0.689 | 0.497 | 0.496 | 0.405 | 0.405 |
| imagenet-o | 0.428 | 0.428 | 0.471 | 0.480 | 0.474 | 0.474 | 0.458 | 0.412 | 0.436 | 0.436 | 0.490 | 0.490 | 0.474 | 0.467 | 0.512 | 0.510 | 0.390 | 0.390 | 0.396 | 0.396 | 0.453 | 0.453 |
| imagenet-r | 0.867 | 0.867 | 0.741 | 0.734 | 0.687 | 0.687 | 0.661 | 0.595 | 0.761 | 0.759 | 0.679 | 0.679 | 0.803 | 0.803 | 0.763 | 0.764 | 0.898 | 0.898 | 0.891 | 0.890 | 0.847 | 0.847 |
| imagenet1k | 0.798 | 0.798 | 0.654 | 0.654 | 0.703 | 0.703 | 0.657 | 0.592 | 0.683 | 0.681 | 0.633 | 0.631 | 0.706 | 0.706 | 0.668 | 0.668 | 0.762 | 0.762 | 0.763 | 0.764 | 0.636 | 0.636 |
| imagenet_sketch | 0.623 | 0.623 | 0.546 | 0.533 | 0.537 | 0.537 | 0.398 | 0.359 | 0.483 | 0.482 | 0.423 | 0.420 | 0.575 | 0.574 | 0.553 | 0.552 | 0.657 | 0.659 | 0.679 | 0.679 | 0.598 | 0.598 |
| imagenetv2 | 0.715 | 0.715 | 0.569 | 0.576 | 0.547 | 0.547 | 0.509 | 0.459 | 0.619 | 0.617 | 0.559 | 0.560 | 0.629 | 0.627 | 0.592 | 0.593 | 0.700 | 0.701 | 0.693 | 0.694 | 0.592 | 0.592 |
| kitti_closest_vehicle_distance | 0.452 | 0.452 | 0.319 | 0.322 | 0.337 | 0.337 | 0.420 | 0.378 | 0.353 | 0.353 | 0.404 | 0.398 | 0.259 | 0.262 | 0.355 | 0.352 | 0.368 | 0.361 | 0.459 | 0.468 | 0.284 | 0.284 |
| mnist | 0.839 | 0.839 | 0.392 | 0.306 | 0.388 | 0.388 | 0.505 | 0.455 | 0.525 | 0.525 | 0.477 | 0.484 | 0.636 | 0.630 | 0.693 | 0.699 | 0.871 | 0.870 | 0.837 | 0.840 | 0.574 | 0.574 |
| pcam | 0.555 | 0.555 | 0.644 | 0.650 | 0.671 | 0.671 | 0.611 | 0.550 | 0.507 | 0.507 | 0.623 | 0.620 | 0.561 | 0.563 | 0.587 | 0.581 | 0.500 | 0.500 | 0.594 | 0.593 | 0.558 | 0.558 |
| pets | 1.000 | 1.000 | 0.885 | 0.881 | 0.914 | 0.914 | 0.836 | 0.753 | 0.888 | 0.888 | 0.870 | 0.868 | 0.912 | 0.912 | 0.904 | 0.905 | 0.923 | 0.923 | 0.943 | 0.943 | 0.995 | 0.995 |
| renderedsst2 | 0.495 | 0.495 | 0.536 | 0.523 | 0.528 | 0.528 | 0.589 | 0.530 | 0.606 | 0.606 | 0.570 | 0.562 | 0.578 | 0.581 | 0.568 | 0.569 | 0.610 | 0.612 | 0.519 | 0.522 | 0.581 | 0.581 |
| resisc45 | 0.626 | 0.626 | 0.612 | 0.610 | 0.649 | 0.649 | 0.507 | 0.457 | 0.588 | 0.588 | 0.541 | 0.541 | 0.634 | 0.633 | 0.612 | 0.614 | 0.644 | 0.634 | 0.646 | 0.647 | 0.619 | 0.619 |
| smallnorb_label_azimuth | 0.045 | 0.045 | 0.050 | 0.050 | 0.055 | 0.055 | 0.059 | 0.053 | 0.052 | 0.052 | 0.062 | 0.059 | 0.056 | 0.059 | 0.062 | 0.061 | 0.058 | 0.060 | 0.048 | 0.048 | 0.056 | 0.056 |
| smallnorb_label_elevation | 0.101 | 0.101 | 0.096 | 0.096 | 0.101 | 0.101 | 0.123 | 0.111 | 0.122 | 0.122 | 0.126 | 0.128 | 0.114 | 0.113 | 0.110 | 0.110 | 0.136 | 0.134 | 0.112 | 0.111 | 0.106 | 0.106 |
| stl10 | 1.000 | 1.000 | 0.952 | 0.964 | 0.860 | 0.860 | 0.911 | 0.820 | 0.983 | 0.983 | 0.971 | 0.971 | 0.978 | 0.977 | 0.964 | 0.963 | 0.941 | 0.926 | 0.980 | 0.980 | 1.000 | 1.000 |
| svhn | 0.577 | 0.577 | 0.254 | 0.251 | 0.245 | 0.245 | 0.146 | 0.132 | 0.348 | 0.348 | 0.144 | 0.137 | 0.531 | 0.524 | 0.415 | 0.411 | 0.670 | 0.670 | 0.544 | 0.537 | 0.500 | 0.500 |
| voc2007 | 0.805 | 0.805 | 0.807 | 0.774 | 0.779 | 0.779 | 0.760 | 0.684 | 0.833 | 0.834 | 0.805 | 0.807 | 0.855 | 0.856 | 0.842 | 0.842 | 0.890 | 0.891 | 0.822 | 0.821 | 0.937 | 0.937 |